\documentclass[11pt, letterpaper, logo, onecolumn, copyright]{arxiv/template_from_google}
\usepackage[T1]{fontenc}

\usepackage{bm} % For bold math symbols like \bm
\usepackage{nicefrac} % For typesetting compact fractions

\usepackage{booktabs} % For professional quality tables
\usepackage{enumitem} % For list customization
\usepackage{fancyhdr}
\usepackage{multirow} % For multi-row cells in tables
\usepackage{tabularx}
\usepackage{multicol}
\usepackage{array}
\usepackage{longtable}

\usepackage{wrapfig} % To wrap text around figures
\usepackage{tocloft} % To control the Table of Contents
\usepackage{fancybox}
\usepackage{authblk} % For author affiliations

\usepackage{graphicx}
\usepackage{caption}
\usepackage[dvipsnames]{xcolor}

\usepackage{xspace} % For intelligent spacing after macros

\usepackage[authoryear, sort&compress, round]{natbib}

\usepackage[dvipsnames]{xcolor} % Make sure xcolor is loaded

\usepackage{hyperref}

\definecolor{darkblue}{rgb}{0.0, 0.0, 0.6}
\definecolor{darkred}{rgb}{0.7, 0.0, 0.0}
\hypersetup{
  pdffitwindow=true,
  pdfstartview={FitH},
  pdfnewwindow=true,
  colorlinks,
  linktocpage=true,
  linkcolor=darkred,
  urlcolor=darkblue,
  citecolor=darkblue
}

\usepackage{amsmath,amssymb}
\usepackage[ruled,vlined,linesnumbered]{algorithm2e}

\usepackage{subfig}
\usepackage{multirow}
\usepackage{multicol}

\usepackage{booktabs}
\usepackage{placeins}
\usepackage{enumitem}

\usepackage{xspace}

\usepackage{fvextra}

\DefineVerbatimEnvironment{PromptBlock}{Verbatim}{
  breaklines=true,
  breakanywhere=true,
  fontsize=\small
}

\usepackage{tcolorbox}

\tcbuselibrary{skins,breakable}

\newcommand{\OpenCode}{\texttt{OpenCode}\xspace}
\newcommand{\AxialCode}{\texttt{AxialCode}\xspace}
\newcommand{\Manage}{\texttt{Manage}\xspace}
\newcommand{\TheoreticalCode}{\texttt{TheoreticalCode}\xspace}

\newcommand{\framework}{\texttt{AutoTraceGT}\xspace}

\newcommand{\SWEGym}{\textsc{Tau-Bench}}
\newcommand{\GoBrowse}{\textsc{Go-Browse}}
\newcommand{\SWEAgent}{\textsc{SWE-Agent}}
\newcommand{\TauBench}{\textsc{Tau-Bench}}
\newcommand{\ALFWorld}{\textsc{ALFWorld}}
\newcommand{\GAIA}{\textsc{GAIA}}
\newcommand{\WebShop}{\textsc{WebShop}}

\let\cite\citep

\title{Using Grounded Theory for Agent Behavior Analysis at Scale}
\reportnumber{} % Leave blank if n/a

\renewcommand{\today}{}

\author[1]{Zhuoran Lu}
\author[2]{Yangyang Yu}
\author[1]{Zhuoyan Li}
\author[3]{Yibo Meng}
\author[4]{Nan Jiang}
\author[3]{Chengxi Zang}
\author[5]{Jie Gao}
\author[5]{Ziang Xiao}

\affil[1]{Purdue University}
\affil[2]{Stevens Institute of Technology}
\affil[3]{Cornell University}
\affil[4]{University of Texas at El Paso}
\affil[5]{Johns Hopkins University}

\renewcommand{\copyrightext}{}
\correspondingauthor={Ziang Xiao, \href{mailto:ziang.xiao@jhu.edu}{ziang.xiao@jhu.edu}\\
\textnormal{Code: \url{https://github.com/ZhuoranLu/Qual-Agent-Behavior-Analysis}}}

\begin{abstract}
  Understanding agent behavior requires methods that scale to thousands of trajectories and surface new patterns in long, often unfamiliar tasks where pre-built classifiers fall short.
  We propose to bring \emph{grounded theory} into agent trajectory analysis: a six-decade-old qualitative method from the social sciences, with a principled saturation criterion and an auditable trail from data to theory. We propose \framework\footnote{\textbf{Auto}mated \textbf{Trace} analysis through \textbf{G}rounded \textbf{T}heory.}, the first multi-agent pipeline that automates grounded theory on agent trajectories: it iteratively performs open, axial, and theoretical coding until saturation, producing a behavioral taxonomy tailored to each task. Across six trajectory corpora, \framework{} produces codebooks that recover $73-91\%$ of the failure modes in human-annotated taxonomies and surface additional patterns that those taxonomies miss. The emergent theoretical narrative aligns with prior expert accounts. Used as a deductive feature space, the codebook outperforms zero-shot and few-shot LLM baselines on downstream failure prediction. These results suggest Grounded Theory offers a scalable analytic tool for ML researchers and agent developers studying what agents actually do.
  \end{abstract}

\begin{document}

\maketitle

  \section{Introduction}
  \begin{figure}[!t]
    \centering
    \includegraphics[width=.6\linewidth]{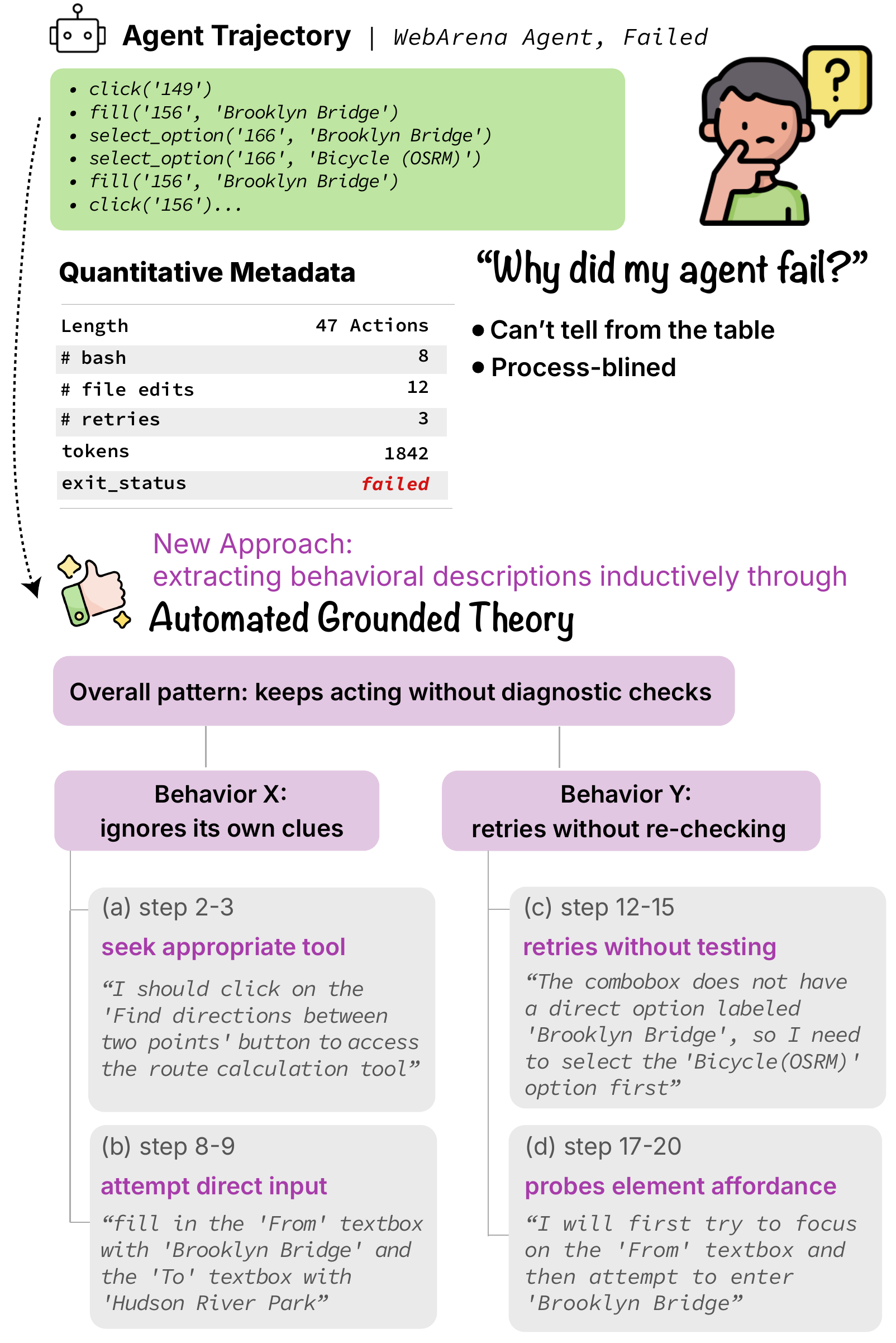}
    \caption{Quantitative metadata records that an agent failed but is \emph{process-blind} to \emph{why}. Automated \emph{grounded theory}~\cite{glaser1967discovery, saldana2021coding} surfaces behavioral patterns from the agent's own reasoning, e.g., \emph{keeps acting without diagnostic checks}, decomposed into sub-behaviors (\emph{ignores its own clues}, \emph{retries without re-checking}) with step-level quoted evidence.}
    \label{fig:motivation}
    % \vspace{-10pt}
\end{figure}

Recent advances in large language models (LLMs) have enabled agents to address increasingly complex tasks, including software engineering~\cite{jimenez2024swe, yang2024swe}, web browsing~\cite{zhou2024webarena, deng2023mind2web}, computer use~\cite{xie2024osworld}, and deep research~\cite{mialon2024gaia}. These systems typically operate through agentic frameworks, in which an LLM interacts with external environments over multi-step trajectories consisting of planning, actions, observations, and revisions. However, stronger task-solving ability does not eliminate brittleness. Agents may still fail after long sequences of locally reasonable decisions, making it difficult to understand which behaviors support successful problem solving and which behaviors lead to failure. This motivates a critical question: \textbf{what do agents actually do while solving tasks, and how can we characterize these behaviors at scale?}

Existing analyses fall into two methodologically constrained regimes. Lightweight quantitative metadata (length, action counts, task success) scales easily but explains little about the underlying problem-solving process~\cite{yang2024swe, jimenez2024swe}. Human analysis is interpretable but expensive, especially for agent trajectories with hundreds of steps~\cite{cemri2026multi, gao2026interpret}. A hybrid approach that builds behavioral classifiers based on pre-defined, often expert-derived, behavioral patterns is too rigid to generalize to novel tasks and emergent agent behaviors. The deeper gap is that the ML community currently lacks a methodology for studying agent behavior that is \emph{scalable} and \emph{generalizable}, a problem the social sciences have long faced and for which they have developed methodological responses.

\textbf{We propose to bring a six-decade-old methodology widely used across qualitative research, \emph{Grounded Theory}, into ML as a new analytic method for agents.} Grounded theory~\cite{glaser1967discovery, charmaz2014constructing} is an inductive qualitative methodology in which theories and categories emerge from the data itself rather than from prior hypotheses: researchers label concrete incidents, group them into categories, and iteratively refine those categories against newly sampled data~\cite{saldana2021coding, williams2019art}. This process terminates at \emph{theoretical saturation}, when new sampling no longer surfaces new structure.
This makes it well-suited to agent trajectory analysis on three counts: it is inductive, and thus open to novel tasks and emergent behaviors; it offers a principled, data-driven stopping criterion via theoretical saturation; and it yields an auditable trail from raw trajectories to theoretical claims. %The catch has historically been bandwidth: grounded-theory studies rarely exceed a few hundred cases. LLMs lift that ceiling: a single call can perform the close reading each coding stage demands, and multi-agent orchestration can carry the iterative round structure across thousands of trajectories.

To this end, we introduce \framework, the first end-to-end pipeline that operationalizes grounded theory for analyzing agent trajectories. Three agents perform layered semantic compression: \OpenCode annotates a single trajectory with descriptive codes for ``what happens at which steps''; \AxialCode groups a batch of coded trajectories into a series of behavioral categories; and \TheoreticalCode integrates these categories into a theoretical account. Running in parallel, \Manage drives the methodology's iterative core: strategic sampling and constant comparison against the running codebook, repeated until saturation.

We evaluate \framework along two axes. For \textbf{process reliability}, we show that across multiple datasets and backbone LLMs, \framework drives codebooks toward saturation and produces codebooks reproducible across independent runs at a level clearly separated from the noise floor of differing data or model inductive biases. For \textbf{artifact quality}, the induced codebooks cover the majority of failure modes in independently constructed human taxonomies while surfacing additional patterns those taxonomies systematically miss; the theoretical narrative independently converges with cascade-of-errors accounts articulated by prior expert analyses; and the codebook can be repurposed as a deductive feature space for failure prediction, outperforming few-shot LLM baselines across multiple backend models.
Our contributions are:
\begin{itemize}[leftmargin=*, itemsep=0pt, topsep=2pt]
    \item \textbf{A new analytic method for ML.} We argue that grounded theory, a six-decade-old method from the social sciences, can be brought into ML as a productive analytic tool for scalable analysis of agent behavior and trajectories.
    \item \textbf{\framework.} We build the first end-to-end pipeline for automated grounded theory, implementing open, axial, and theoretical coding as role-specialized agents coordinated by a codebook manager, with every intermediate artifact machine-readable and auditable.
    \item \textbf{Empirical validation.} On 7{,}500+ trajectories across six datasets and four backbone LLMs, we show that \framework reaches saturation and produces reproducible codebooks, covers the majority of human-taxonomy failure modes, recovers prior expert theoretical accounts, and yields a deductive feature space for failure prediction.
\end{itemize}

  \section{Related Work}

\begin{figure*}[t]
    \centering
    \includegraphics[width=\textwidth]{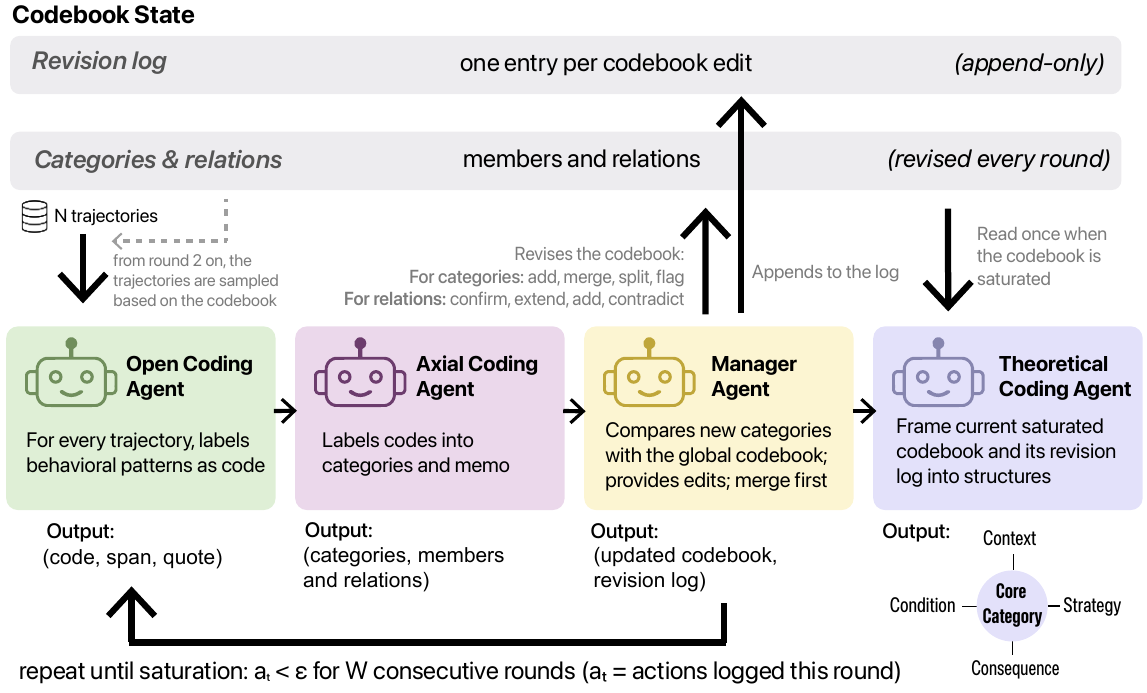}
    \caption{Overview of our multi-agent qualitative coding pipeline. }
    \label{fig:pipeline}
\end{figure*}

\textbf{Grounded theory} is an inductive qualitative methodology~\cite{glaser1967discovery, charmaz2014constructing} widely used across the social sciences, with close analogues in thematic analysis~\cite{braun2006using} and content analysis~\cite{krippendorff2018content}. Its core idea is to build theories \emph{grounded} in data rather than test pre-existing hypotheses, which suits phenomena where established theories do not apply or new insights are sought.
The method is realized through qualitative coding~\cite{saldana2021coding, williams2019art}, in which the analyst proceeds in three layered stages. \emph{Open coding} labels salient meanings in each document with descriptive tags. \emph{Axial coding} links these tags across documents (e.g., by cause, condition, consequence) and groups them into higher-level categories. \emph{Theoretical} (or \emph{selective}) \emph{coding} integrates the saturated categories into a coherent account of the phenomenon. Because every category and claim stays linked to the incidents that produced it, grounded theory yields human-readable patterns with an auditable trail back to the raw data, a natural fit for analyzing open-ended agent trajectories at scale.

\noindent\textbf{LLM-assisted Qualitative Analysis.} LLMs have been used to scale qualitative coding, including thematic analysis on interview transcripts~\cite{xiao2023supporting,de2024performing}, hierarchical inductive coding~\cite{zhong2025hicode, gao2024collabcoder}, multi-agent thematic analysis~\cite{yi2025auto, lin2026agentaspeerdebriefer}, and dedicated grounded-theory pipelines~\cite{ubellacker2024academiaos,pi2025logos}. We focus on the last category, where two gaps remain. First, grounded theory's multi-stage structure is rarely preserved end-to-end: AcademiaOS~\cite{ubellacker2024academiaos} offers limited cross-stage orchestration, and LOGOS~\cite{pi2025logos} replaces axial and selective coding with semantic clustering, foregoing the role-differentiated analytic stance the methodology relies on. Second, \emph{theoretical saturation} (grounded theory's own termination criterion) is not empirically verified by these pipelines, which instead stop after a fixed number of iterations; whether the produced codebook reflects conceptual breadth or an arbitrary stop is left unclear. \framework addresses both: it implements the three stages as role-specialized agents with an explicit cross-batch manager, and verifies saturation through codebook convergence across iterations.

\noindent\textbf{Agent Behavior Analysis.}
LLM-based agents are typically benchmarked by task-completion success on suites such as SWE-bench ~\cite{jimenez2024swe} and WebArena ~\cite{zhou2024webarena} which report whether trajectories succeed but reveal little about why they fail. A growing line of work addresses this through structured failure analysis: AgentErrorTaxonomy ~\cite{zhu2025llm} decomposes failures into five operational modules, MAST~\cite{cemri2026multi} applies grounded theory by hand to 150+ multi-agent traces and produces a 14-mode taxonomy, and others target subproblems such as tool-parameter failures~\cite{xiong2025butterfly} or rubric-based trajectory verification ~\cite{raghavendra2026agentic}. Across this literature, taxonomies are built either by researchers iterating on small samples or by LLMs prompted to enumerate failure modes in one pass—neither route makes the inductive process auditable or guarantees a verifiable stopping criterion. \framework operationalizes the full Grounded Theory loop algorithmically and scales it to thousands of trajectories, with theoretical saturation verified empirically.
  \section{Multi-Agent Framework}
  % \jie{lack of an overview sentence here}
% \paragraph{Grounded theory.} Grounded theory~\cite{glaser1998grounded, saldana2021coding} inductively builds theory from data through three coding stages: \emph{open coding} labels discrete incidents with conceptual codes, \emph{axial coding} groups codes into categories with typed relations, and \emph{theoretical coding} integrates categories around a core category. These stages are driven by an iterative \emph{round structure}: \emph{theoretical sampling} draws each round of data informed by what has emerged so far; the round is open- then axial-coded; and \emph{constant comparison} folds it into a running codebook. The loop terminates at \emph{theoretical saturation}, when new data no longer yield new categories. In what follows, we describe how \framework operationalizes these principles at the scale of thousands of agent trajectories. 
% \jie{TODO: put this one into related work. I think here should start from a clean overview}

\begin{algorithm*}[!ht]
\caption{ Automated Grounded Theory for Agent Trajectories}
\label{alg:multi_agent_coding}
\KwIn{Pool of~$N$ trajectories; batch size~$B$; sampling policy $\pi$; saturation threshold~$\epsilon$; stability window~$W$}
\KwOut{Saturated codebook $K$, theoretical account $(c^*, \mathcal{N})$}
Initialize $K \leftarrow \emptyset$;\quad $t \leftarrow 0$\;

\Repeat(\tcp*[f]{\textcolor{blue}{Saturate after continuous $W$ rounds with additions $< \epsilon$}})
{$t \geq W$ \textbf{ and } $a_j < \epsilon,\ \text{for all } j \in \{t-W+1,\ldots,t\}$}{
    $t \leftarrow t + 1$\;
    Draw a batch of $B$ trajectories from the pool via $\pi$, conditioned on $K$\;
    \ForEach{trajectory in the batch}{
        Run $\OpenCode$ chunk by chunk, carrying a segment memo\;
    }
    $A \leftarrow \AxialCode(\text{coded batch})$;
    $K \leftarrow \Manage(K, A)$;
    $a_t \leftarrow$ \#\textsc{add} actions logged this round\;
}
$(c^*, \mathcal{N}) \leftarrow \TheoreticalCode(K)$\;
\Return{$K,\ (c^*, \mathcal{N})$}
\end{algorithm*}

\framework takes a corpus of 
$N$ agent trajectories and conducts grounded theory analysis over it. The framework comprises four agents operating at three scales: \OpenCode works on a single trajectory, \AxialCode on a batch of trajectories, and \TheoreticalCode on the full corpus to synthesize the emergent theory; orthogonal to these, \Manage maintains analytic state and continuity across batches. Figure~\ref{fig:pipeline} presents the framework overview, and we describe each term below.

\paragraph{Open coding \textit{(per trajectory)}.} The open-coding agent, $\OpenCode: \text{Trajectory} \to \text{Records}$, labels behavioral incidents on a trajectory with conceptual codes. Each output record takes the form $(\text{code},\, \text{span},\, \text{quote})$: a 2--5 word conceptual code describing what happens, a sub-step span indicating where it happens, a short verbatim quote as the evidence. The number of codes per trajectory is chosen adaptively by the agent. We instruct it to abstract behavioral patterns rather than restate surface-level actions to use the conceptual code captures recurrent modes of conduct (e.g., \emph{diagnose environment constraints}, \emph{persist through validation hurdles}) instead of literal tool calls. Operationally, to keep each LLM call within context, we partition the message sequence $\mathbf{m}$ into chunks and run $\OpenCode$ chunk by chunk, carrying a segment memo across calls to preserve analytic continuity (see Algorithm~\ref{alg:multi_agent_coding}). Coding is sequential within a trajectory but parallelizable across trajectories.

\paragraph{Axial coding \textit{(per sampling round)}.} Following the iterative structure of grounded theory, analysis proceeds round by round: each round is a batch of open-coded trajectories drawn under a sampling policy and fed to the axial-coding agent. $\AxialCode: \text{CodedTrajectories} \to \text{Categories} \times \text{Relations}$ groups the round's conceptual codes into categories and the typed relations between them, and emits an axial memo $\mu^{\text{ax}}$ summarizing the round. Each induced category carries a textual definition, the set of code records supporting it, and a status tag $\sigma \in \{\textsc{succ}, \textsc{fail}, \textsc{both}\}$ derived by aggregating the terminal status of each member's source trajectory. The membership pointers preserve an explicit link from each category back to its supporting episodes and source trajectories.

\paragraph{Codebook management \textit{(constant comparison across rounds)}.} Grounded theory's practice of continuously revising codes against incoming data is realized explicitly as the codebook-manager $\Manage: \text{Codebook} \times \text{RoundStructure} \to \text{Codebook}$, which reconciles each round's axial output against the running codebook. We maintain a versioned codebook state $K_t = (\mathcal{C}_t, \mathcal{R}_t, L_t)$ with a revision log $L_t$; on each new round, $\Manage$ produces $K_{t+1}$ by selecting for each new category an action from $\{\texttt{add}, \texttt{merge}, \texttt{split}, \texttt{flag}\}$ and for each new relation from $\{\texttt{confirm}, \texttt{extend}, \texttt{add}, \texttt{contradict}\}$, with every action appended to $L_{t+1}$. We adopt a merge-first policy, aligning new categories to existing structure whenever a compatible match exists. Define the number of newly added categories in round $t$ as
$
a_t = |\{\ell \in L_t \setminus L_{t-1}: \ell.\text{action} = \texttt{add}\}|.
$
We operationalize \emph{theoretical saturation} as the criterion $a_t < \epsilon$ and $a_{t-1} < \epsilon$ for two consecutive rounds, and define $T$ as the first round at which the criterion is met. The resulting $L_T$ provides an auditable history of theory evolution.

\paragraph{Theoretical coding \textit{(global level)}.} After saturation, the codebook stabilizes at $K_T$, and the theoretical-coding agent, $\TheoreticalCode: \text{Codebook} \to \text{CoreCategory} \times \text{Narrative}$, produces $(c^*, \mathcal{N})$ where $c^* \in \mathcal{C}_T$ is the core category and $\mathcal{N}$ is a narrative account that arranges the remaining categories around $c^*$ as conditions, contexts, strategies, and consequences. Crucially, this stage depends only on $K_T$ and the revision log $L_T$, introducing no new local evidence: it consolidates the global structure produced by earlier agents while preserving unresolved tensions and ambiguous cases.

% \paragraph{Theoretical coding \textit{(global level)}.} After $T$ batches the codebook stabilizes at $K_T$, and the theoretical-coding agent has interface $\TheoreticalCode: \text{Codebook} \to \text{CoreCategory} \times \text{Roles} \times \text{Narrative}$, producing $(c^*, \rho, \mathcal{N})$ where $c^* \in \mathcal{C}_T$ is the core category, $\rho$ assigns each remaining category a role relative to $c^*$ (e.g., condition, context, strategy, consequence), and $\mathcal{N}$ is a narrative theoretical account of agent behavior. Crucially, this stage depends only on $K_T$ and the revision log $L_T$, introducing no new local evidence: it consolidates the global structure produced by earlier agents while preserving unresolved tensions and ambiguous cases. 

\begin{figure*}[t]
    \centering
    \includegraphics[width=\textwidth]{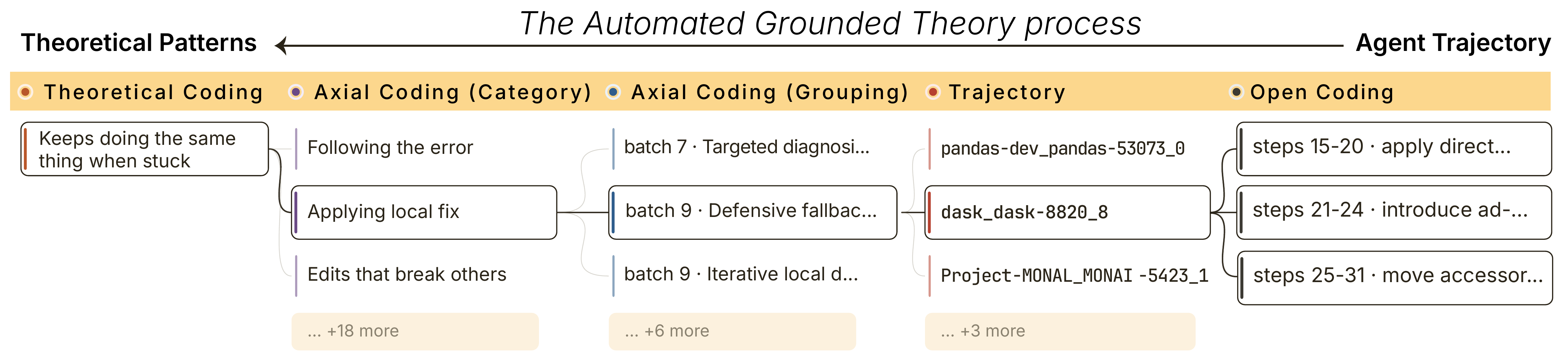}
    \caption{An example of \framework's coding process. \OpenCode annotates a raw trajectory with conceptual open codes; \AxialCode groups open codes across batches into categories and typed relations; \TheoreticalCode integrates the saturated codebook into a theoretical account. Every code, category, and claim points back to its supporting evidence, making each step auditable.}
    \label{fig:audit-example}
\end{figure*}
% The codebook $K_T$ serves as the structured artifact consumed by downstream tasks such as the skill-induction-based failure detection in Table~\ref{tab:failure_detection_multi_endpoint}, while $(c^*, \rho, \mathcal{N})$ provides a human-interpretable theoretical account.
% After the codebook stabilizes, a \textbf{selective-coding agent} consumes the final codebook together with the accumulated revision log. It identifies a core category, assigns each remaining category a role relative to that core, and generates an integrative theoretical account of agent behavior across trajectories. Rather than introducing new local evidence, this stage consolidates the global structure produced by earlier agents into an explicit explanatory narrative while preserving unresolved tensions and ambiguous cases.

% All intermediate artifacts are machine-readable, including episode-level code records, axial summaries, codebook revisions, and the final selective-coding output. This design makes the analysis process fully traceable from raw trajectory messages to final theoretical claims, and supports inspection, debugging, and re-analysis.

  \section{Experiments}

In this section, we conduct a comprehensive evaluation of \framework on multiple agent trajectory datasets, primarily to answer the following two research questions:
\begin{itemize}[leftmargin=*, itemsep=2pt, topsep=2pt, parsep=0pt]

    \item \textbf{RQ1 -- Process Reliability:} Does \framework instantiate grounded theory methodology faithfully and stably?

    \item \textbf{RQ2 -- Artifact Quality:} Are \framework outputs, including codebooks and theoretical accounts, grounded, interpretable, and useful?

\end{itemize}

To answer these two research questions, we used two types of corpora (see details in Appendix~\ref{appendix:corpora}):

\noindent\textbf{Trajectories with outcome labels.}
These corpora span three environments covering different task and reasoning 
profiles: \SWEGym~\cite{yao2024tau}, \GoBrowse~\cite{gandhi2025go}, and \SWEAgent~\cite{yang2024swe}. From each environment, we sample 2{,}000 trajectories. Each trajectory carries only an environment-provided outcome label 
(i.e., success or failure). 

\noindent\textbf{Trajectories with annotated failure reasoning.}
These corpora consist of agent trajectories from three single-agent environments spanning diverse interaction patterns and reasoning demands:
\ALFWorld, \GAIA, and \WebShop. Each trajectory
is paired with expert annotations identifying failure types and their 
causal reasoning~\cite{zhu2025llm}.

\subsection{RQ1: Process Reliability}

\begin{figure*}[t]
    \centering
    \includegraphics[width=1.0\textwidth]{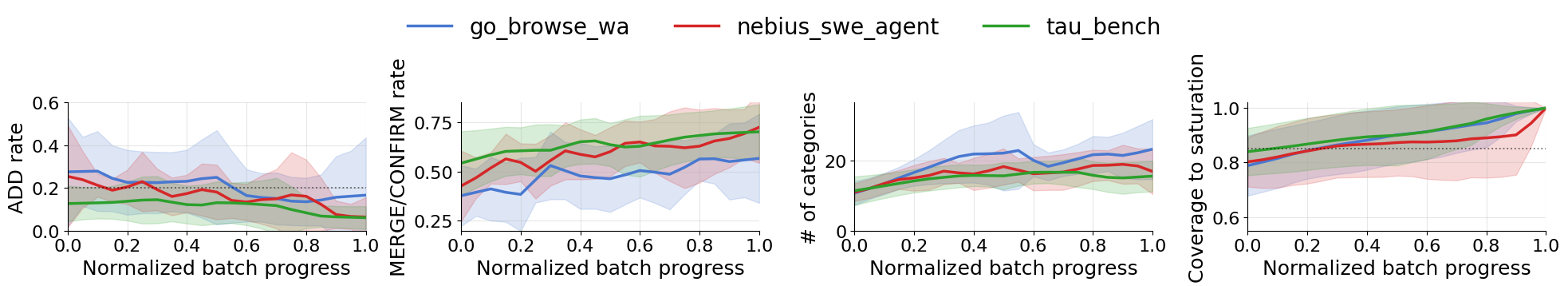}
    \caption{Saturation diagnostics for \framework. All four panels share an
    $x$-axis of \emph{normalized coding progress}: because different runs use
    different numbers of rounds, each run's timeline is rescaled to $[0, 1]$ so
    that runs of varying lengths can be overlaid. \textbf{(a)} Per-round
    fraction of \Manage actions that are \textsc{add}. \textbf{(b)} Per-round
    fraction of \Manage actions that are \textsc{merge} or \textsc{confirm}.
    \textbf{(c)} Total number of categories in the running codebook.
    \textbf{(d)} Cosine similarity between the running codebook and the
    terminal saturated codebook. \textsc{add} tapers while
    \textsc{merge}/\textsc{confirm} rise (a,~b); the codebook stabilizes in
    size (c) and content (d), thus indicating saturation.}
    \label{fig:saturation}
\end{figure*}

\begin{figure*}[t]
  \centering
  \subfloat[Within-config replicate coverage vs. cross-config null \label{fig:coverage-stability:a}]{
    \includegraphics[width=\columnwidth]{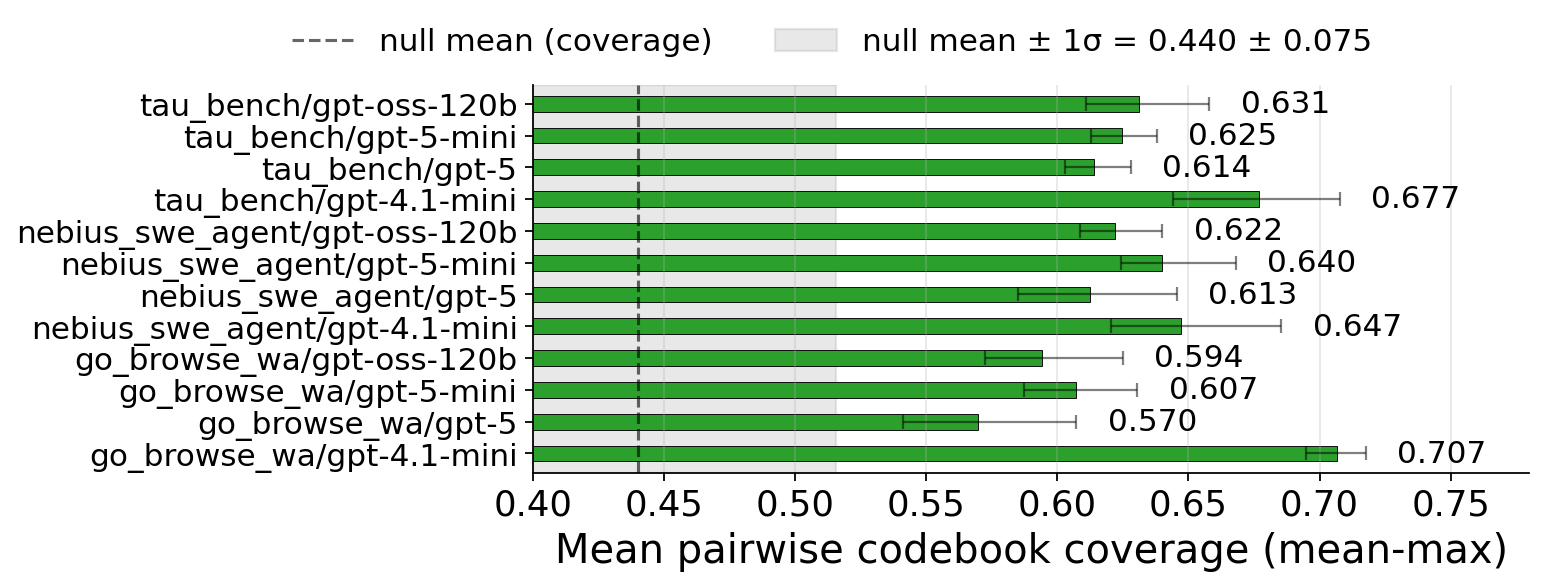}
  }
  \hfill
  \subfloat[Cross-LLM same-dataset coverage vs. null baselines\label{fig:coverage-stability:b}]{
    \includegraphics[width=\columnwidth]{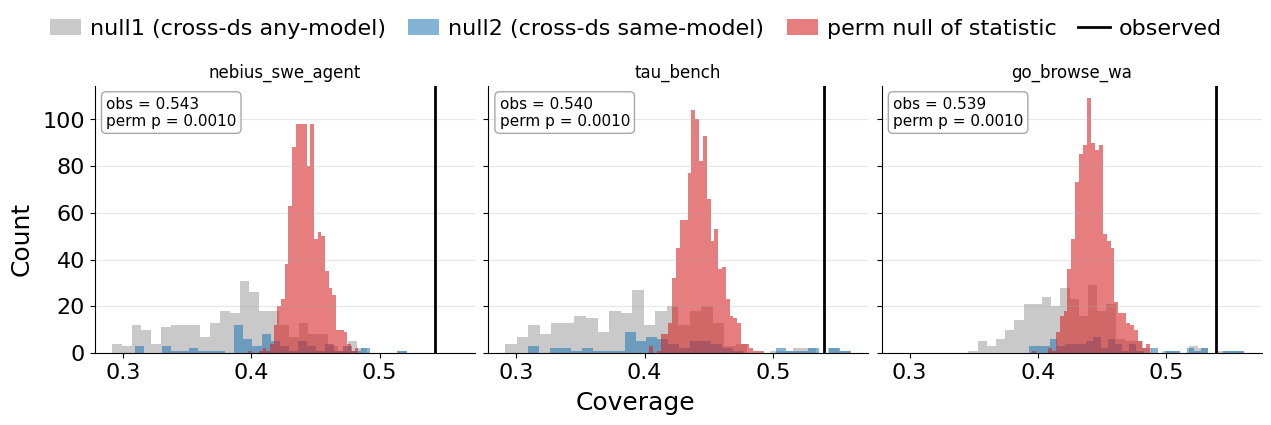}
  }

\caption{\textbf{(a)} Pairwise coverage between final codebooks from replicate runs under the same dataset--model configuration. 
\textbf{(b)} Cross-LLM coverage within each dataset compared against cross-dataset and permutation-based null baselines. 
Full null statistics and permutation details are reported in Appendix~\ref{app:coverage-stability}.}
\label{fig:coverage-stability}

\end{figure*}

\noindent\textbf{\framework drives codebooks toward saturation.} Figure~\ref{fig:saturation} reports four diagnostics over normalized coding progress: two on the operations \Manage issues each round (a, b), and two on the resulting state of the codebook (c, d). As coding progresses, \textsc{add} actions taper off while \textsc{merge} and \textsc{confirm} take over (a, b), as new patterns are increasingly absorbed into existing categories rather than spawning new ones---the behavioral signature of theoretical saturation, where additional data ceases to yield new categories~\cite{saunders2018saturation}. This taper is also the algorithmic stopping signal: \framework terminates when $a_t < \epsilon$ for consecutive rounds (Algorithm~\ref{alg:multi_agent_coding}). The codebook itself stabilizes accordingly: category count plateaus at a dataset-specific value and cosine similarity to the terminal codebook approaches $1$ (c, d).

\noindent \emph{Within-configuration replicate stability.}
We first test whether \framework produces stable codebooks under repeated runs of the same configuration. For each dataset--model configuration, we run \framework three times on disjoint trajectory subsets and compare the resulting saturated codebooks. Codebooks produced within the same configuration are significantly more similar than codebooks produced across configurations (Mann--Whitney $U$, $p < 10^{-20}$). Moreover, all configuration-level means exceed a data-derived reproducibility threshold estimated from cross-configuration comparisons (Figure~\ref{fig:coverage-stability:a}). Full construction and distributional statistics are reported in Appendix~\ref{app:coverage-stability}.

\noindent\emph{Cross-LLM stability on the same dataset.}
We next test whether the induced codebooks reflect trajectory-specific structure rather than backend-model priors. Holding the dataset fixed, we compare codebooks produced by different backend LLMs and contrast them against null comparisons that break the dataset correspondence. Across all datasets, cross-LLM coverage remains above the null baselines, with permutation tests significant for every dataset ($p < 0.001$; Figure~\ref{fig:coverage-stability:b}). This suggests that \framework recovers dataset-specific behavioral structure that is stable across backend models. Details of the null construction and permutation procedure are reported in Appendix~\ref{app:coverage-stability}.

\subsection{RQ2: Artifact Quality}
\label{sec:rq2}
We evaluate whether \framework's outputs are grounded, interpretable, and useful. This section presents two complementary analyses: coverage against human failure annotations (Section~\ref{sec:coverage}), and downstream failure detection (Section~\ref{sec:detection}).

\subsubsection{Do \framework's findings echo expert analyses of trajectories?}
\label{sec:coverage}

\paragraph{Coverage against Human Taxonomy}
We compare \framework's inductively derived codebook against an independently constructed human taxonomy of agent failure modes from prior work, which manually analyzed trajectories on three benchmarks (\ALFWorld~\cite{shridhar2020alfworld}, \GAIA~\cite{mialon2024gaia}, \WebShop~\cite{yao2022webshop}), where
domain experts inductively coded $\sim$500 trajectories per benchmark 
to produce a closed list of failure types, denoted $\mathcal{H}$, 
along with per-trajectory free-text reasoning $r_t$ explaining the 
specific failure mechanism. We run \framework with GPT-5-mini on 
the same trajectories, producing codebook $\mathcal{C}$. The comparison 
between $\mathcal{C}$ and $\mathcal{H}$ measures how much two 
independent inductive analyses (one human, one computational) converge
on the same behavioral patterns. For each trajectory, we use GPT-5 as an LLM judge to 
determine whether any candidate covers $r_t$ (see 
Appendix~\ref{appendix:coverage:protocol} for details). We report the coverage precision and recall between the two codebooks, and the fraction of trajectories whose human-written reasoning is covered by $\mathcal{C}$.
% \begin{itemize}[leftmargin=*, itemsep=0pt, topsep=2pt]
%   \item $C \!\to\! H$: fraction of human-induced failure modes in 
%   $\mathcal{H}$ is covered by $\mathcal{C}$.
%   \item $H \!\to\! C$: fraction of \framework-induced failure modes in 
%   $\mathcal{C}$ is covered by $\mathcal{H}$.
%   \item Match: fraction of trajectories whose human-written reasoning is covered by $\mathcal{C}$.
% \end{itemize}

Table~\ref{tab:coverage-gpt5mini} reports mutual coverage across 
datasets. The codebook covers $\geq$70\% of human failure modes 
on all three datasets (peaking at 90.9\% on \WebShop), and matches
58--88\% of individual trajectories' reasonings. 

\begin{table}[!t]
\centering
\begin{tabular}{lrrrrr}
\toprule
Dataset & $|H|$ & $|C|$ & Recall & Precision & Match \\
\midrule
\ALFWorld & 15 & 20 & 75.0 & 60.0 & 82.4 \\
\GAIA     & 11 & 17 & 73.7 & 63.2 & 58.0 \\
\WebShop  & 10 & 13 & 90.9 & 88.9 & 87.9 \\
\bottomrule
\end{tabular}
% \vspace{-0.8em}
\caption{Mutual coverage between the \framework codebook and human failure annotations. 
$|\mathcal{H}|$ / $|\mathcal{C}|$: human /\framework codebook categories. Metrics are defined in Section~\ref{sec:coverage}.} \label{tab:coverage-gpt5mini}
% \vspace{-0.8em}
\end{table}

% Declare the compact GLM table here so it is placed one page before the
% full-width failure-prediction table.
\iffalse
\begin{table}[!t]
\centering
\resizebox{\columnwidth}{!}{%
\begin{tabular}{llr}
\toprule
Theory & GLM features & $\beta$ \\
\midrule
\textbf{T1}
  & $\times$ Uses sweeping workaround.
  & $+4.10$ \\
 \SWEAgent & $\checkmark$ Builds minimal reproducer after reconnaissance.
  & $-1.82$ \\
\midrule
\textbf{T2}
  & $\checkmark$ Clicks visible UI affordance.
  & $-2.59$ \\
\GoBrowse  & $\times$ Clicks UI; closes without verification.
  & $+3.28$ \\
\midrule
\textbf{T3}
  & $\checkmark$ Executes after consent and reconciliation.
  & $-0.99$ \\
 \TauBench & $\times$ Finds payment error late; escalates.
  & $+1.64$ \\
\bottomrule
\end{tabular}%
}
\caption{GLM support for \textbf{T1}--\textbf{T3}. $\checkmark$/$\times$: associated with success/failure.}
\label{tab:lr_evidence}
\end{table}
\fi

Notably, the recall can exceed precision for $\mathcal{H}$ as the reference, indicating that \framework surfaces 
categories absent from the predefined human-induced taxonomy. These uncovered modes are not random gaps but systematic limitations of pre-specified failure 
taxonomies. We identified one example code per benchmark (full definitions in 
App.~\ref{appendix:emergence}) that was not covered by $\mathcal{H}$. First, granularity: \ALFWorld's
\emph{noncompliant inaction}, in which the agent produces no 
admissible action when one is required, is distinguishable from human-authored
\texttt{invalid\_action} and \texttt{impossible\_action} (both of 
which presuppose an attempted action) only by the complete absence 
of action, a distinction that pre-specified taxonomies often 
collapse. Second, cross-module structure: \GAIA's
\emph{signaled-but-unrealized shifts}, in which the agent announces 
a pivot in plan that is never enacted, is a planning--execution 
decoupling that no single cognitive-module label can express because 
the failure spans the boundary between modules. Third, catch-all labeling: the generic inefficient\_plan label collapses WebShop's mode oscillation without synthesis, rapid toggling between paging and resets without feedback integration, into a single bucket alongside distinct inefficiencies that call for different remediations.

\paragraph{Theoretical convergence with human-authored analyses.}
The theoretical coding stage of \framework produces a narrative account of failure that converges in substance with---while remaining complementary in framing to---the cascade-of-errors view articulated by the same prior work~\cite{zhu2025llm}, which theorized that early errors propagate downstream as upstream cognitive modules distort subsequent planning. Without access to their analysis, \framework's theoretical coding recovers the same mechanism at the behavioral surface: across all three benchmarks, the core category names an action stream that has stopped integrating environmental feedback, preserving and re-enacting the original error (\textit{feedback-decoupled control} on \ALFWorld, \textit{persistent repetition without adaptation} on \GAIA, and \textit{ritualized non-diagnostic search} on \WebShop). The two accounts identify the same underlying phenomenon, an agent locked out of adaptive feedback integration, but frame it at different levels of abstraction: prior work attributes the cascade to specific upstream cognitive modules, whereas our action-grounded narrative specifies what this lockout looks like as observable behavior, without positing internal modules. We read this level-complementarity as a strength: two independent inductive analyses, working from the same trajectories but at different framings, arriving at the same mechanism is exactly the cross-validation grounded theory aims for.

\begin{table}[!b]
\centering
\begin{tabular}{llr}
\toprule
Theory & GLM features & $\beta$ \\
\midrule
\textbf{T1}
  & $\times$ Uses sweeping workaround.
  & $+4.10$ \\
 \SWEAgent & $\checkmark$ Builds minimal reproducer after reconnaissance.
  & $-1.82$ \\
\midrule
\textbf{T2}
  & $\checkmark$ Clicks visible UI affordance.
  & $-2.59$ \\
\GoBrowse  & $\times$ Clicks UI; closes without verification.
  & $+3.28$ \\
\midrule
\textbf{T3}
  & $\checkmark$ Executes after consent and reconciliation.
  & $-0.99$ \\
 \TauBench & $\times$ Finds payment error late; escalates.
  & $+1.64$ \\
\bottomrule
\end{tabular}
\caption{GLM support for \textbf{T1}--\textbf{T3}. $\checkmark$/$\times$: associated with success/failure.}
\label{tab:lr_evidence}
\end{table}

\subsubsection{How does \framework characterize agent behavior in successful and failed trajectories?}
\label{sec:detection}

% Declare this full-width float early so it can be placed at the top of the
% page before the Conclusion instead of being deferred to the Conclusion page.
\begin{table*}[!t]
\centering
  \resizebox{\linewidth}{!}{%
\begin{tabular}{ll|cc|cc|cc}
\toprule
& & \multicolumn{2}{c|}{\TauBench} & \multicolumn{2}{c|}{\GoBrowse} & \multicolumn{2}{c}{\SWEAgent} \\
\cmidrule(lr){3-4} \cmidrule(lr){5-6} \cmidrule(lr){7-8}
\textbf{Model} & \textbf{Method} & MCC & ROC AUC & MCC & ROC AUC & MCC & ROC AUC \\
\midrule
\multirow{4}{*}{GPT-4.1-mini}
 & Few-shot & $0.038$ & $0.516$ & $0.220$ & $0.620$ & $0.243$ & $0.720$ \\
 & Few-shot Codebook Feature Engineering & $0.245$ & $0.656$ & $0.325$ & $0.738$ & $0.274$ & $0.712$ \\
 & \framework-Codebook Feature Engineering$^\dagger$ & $0.153$ & $0.612$ & $0.282$ & $0.706$ & $\mathbf{0.314}\uparrow$ & $\mathbf{0.765}\uparrow$ \\
 & \framework Complementary Feature Engineering$^\dagger$ & $0.227$ & $0.656$ & $\mathbf{0.379}\uparrow$ & $\mathbf{0.773}\uparrow$ & $\mathbf{0.343}\uparrow$ & $\mathbf{0.780}\uparrow$ \\
\midrule
\multirow{4}{*}{GPT-5}
 & Few-shot & $0.061$ & $0.550$ & $0.358$ & $0.679$ & $0.351$ & $0.804$ \\
 & Few-shot Codebook Feature Engineering & $0.202$ & $0.622$ & $0.343$ & $0.736$ & $0.320$ & $0.708$ \\
 & \framework Codebook Feature Engineering$^\dagger$ & $0.120$ & $0.596$ & $\mathbf{0.498}\uparrow$ & $\mathbf{0.807}\uparrow$ & $0.326$ & $0.769$ \\
 & \framework Complementary Feature Engineering$^\dagger$ & $\mathbf{0.206}\uparrow$ & $\mathbf{0.669}\uparrow$ & $\mathbf{0.499}\uparrow$ & $\mathbf{0.828}\uparrow$ & $\mathbf{0.372}\uparrow$ & $0.785$ \\
\midrule
\multirow{4}{*}{GPT-5-mini}
 & Few-shot & $0.061$ & $0.531$ & $0.291$ & $0.635$ & $0.222$ & $0.702$ \\
 & Few-shot Codebook Feature Engineering & $0.257$ & $0.663$ & $0.370$ & $0.749$ & $0.272$ & $0.724$ \\
 & \framework Codebook Feature Engineering$^\dagger$ & $0.151$ & $0.624$ & $\mathbf{0.374}\uparrow$ & $\mathbf{0.751}\uparrow$ & $\mathbf{0.350}\uparrow$ & $\mathbf{0.770}\uparrow$ \\
 & \framework Complementary Feature Engineering$^\dagger$ & $\mathbf{0.306}\uparrow$ & $\mathbf{0.699}\uparrow$ & $\mathbf{0.425}\uparrow$ & $\mathbf{0.788}\uparrow$ & $\mathbf{0.318}\uparrow$ & $\mathbf{0.772}\uparrow$ \\
\midrule
\multirow{4}{*}{GPT-OSS-120B}
 & Few-shot & $0.028$ & $0.516$ & $0.239$ & $0.609$ & $0.262$ & $0.713$ \\
 & Few-shot Codebook Feature Engineering & $0.211$ & $0.654$ & $0.244$ & $0.638$ & $0.126$ & $0.607$ \\
 & \framework Codebook Feature Engineering$^\dagger$ & $0.190$ & $0.615$ & $\mathbf{0.383}\uparrow$ & $\mathbf{0.755}\uparrow$ & $\mathbf{0.278}\uparrow$ & $0.704$ \\
 & \framework Complementary Feature Engineering$^\dagger$ & $\mathbf{0.289}\uparrow$ & $\mathbf{0.672}\uparrow$ & $\mathbf{0.390}\uparrow$ & $\mathbf{0.744}\uparrow$ & $0.240$ & $\mathbf{0.726}\uparrow$ \\
 \bottomrule
\end{tabular}%
}
\caption{MCC and ROC AUC on test sets. Codebook Feature Engineering = AutoML over codebook annotations.  $\mathbf{bold\uparrow}$ = exceed both baselines. $^\dagger$ = benefits from adding the \framework codebook. See App.~\ref{appendix:implementation} for details.}
\label{tab:failure_pred}
\end{table*}

\framework produces codebooks and derived theories that characterize agent trajectories. We use them in two ways: to interpret behavioral patterns in trajectories, and to operationalize the patterns for downstream failure detection.

\paragraph{Interpreting behavioral patterns in trajectories.}

Theoretical coding allows us to derive behavior-level explanations of why trajectories succeed or fail. We illustrate this with one representative theoretical finding from each benchmark.

In \SWEAgent, we observe (\textbf{T1}) that successful trajectories tend to ground a proposed fix in executable evidence before modifying the code. Concretely, they first construct or identify a runnable reproducer, connect the observed failure to a specific code region, and then make a targeted edit. Failed trajectories, by contrast, often act without this grounding: they rely on broad rewrites, plausible but unchecked assumptions, or superficial tests that do not reproduce the original failure. In \GoBrowse, we find (\textbf{T2}) that clicking visible UI affordances is not inherently predictive of success or failure. What separates resolved from failed trajectories is whether the agent inserts a substantive verification step between the action and the closure that declares the task complete. In \TauBench, we highlight (\textbf{T3}) that successful trajectories resolve the relevant task state before taking irreversible actions. In these trajectories, the agent aligns the user's identity, retrieved records, constraints, and payment status into a coherent action plan before execution. Failed trajectories often postpone this alignment until after they have already committed to an action, which leads to inconsistent state assumptions, failed execution, or escalation to a human handoff.

\label{sec:annotation}

In addition to these narrative explanations, \framework provides an exploratory quantitative layer for triangulating the derived theories. We repurpose the codebook as a deductive feature schema and construct two types of trajectory-level features. First, for each codebook category, we define a \textit{presence} feature indicating whether the trajectory contains concrete steps that instantiate the category definition. Second, for each relationship empirically confirmed during axial coding, we define \textit{co-occurrence} indicating whether the related categories appear together in the same trajectory. Each trajectory is therefore represented as a fixed-length binary vector consisting of presence and co-occurrence features defined by the codebook.

We use these vectors as predictors and the trajectory outcome as the dependent variable, with failure coded as 1 and success coded as 0. We then fit an interpretive generalized linear model (GLM) whose coefficients can be read alongside the qualitative theory, with positive/negative coefficients indicating features associated with failure/success.

We use GPT-5-mini codebooks to annotate category presence and co-occurrence across all trajectories from the three datasets, and fit the GLM described above (see Appendix~\ref{tab:lr_fit} for fit diagnostics). Table~\ref{tab:lr_evidence} further shows how the annotated features empirically support the derived theories. For example, in \GoBrowse, clicking visible UI affordances alone is associated with successful trajectories ($\beta = -2.59$), suggesting that UI clicking is not inherently a failure signal. However, clicking a UI affordance and then declaring completion without substantive verification substantially increases the likelihood of failure ($\beta = +3.28$). This contrast supports \textbf{T2} derived from \GoBrowse: the decisive behavioral pattern is not whether to click, but whether the agent verifies the task state before closure.

\paragraph{Codebook for deductive annotation supports failure detection.}
The preceding sections show that \framework artifacts characterize agent failures interpretively. We now examine whether the artifacts can also support predictive use. Given only a partial trajectory, we ask whether a model can use the behavioral signals encoded in the codebook to forecast eventual failure.

To test this, we use codebook annotations as features for failure classification~\cite{li2026human}. Following Section~\ref{sec:annotation}, each trajectory prefix is represented as a fixed-length vector of presence and co-occurrence features over codebook categories. We feed this vector to FLAML~\cite{wang2021flaml} AutoML to train a classifier. We compare four variants. The first uses the \framework codebook. The second uses a baseline codebook induced through few-shot prompting, where the prompt asks the LLM to identify concrete behavioral categories from example trajectories. The third uses the union of features from both codebooks. The fourth is a few-shot baseline in which an LLM directly predicts failure without structured codebook features. For the three AutoML variants, we keep the downstream annotation and modeling procedure identical, changing only the source of the codebook features. Table~\ref{tab:failure_pred} reports the MCC and ROC AUC to measure the performance of different approaches on this imbalanced detection problem. The results show that the \framework codebook provides useful predictive signal, though not always as a standalone replacement for baselines.
\framework features outperform all baselines on some datasets, while their combination with few-shot codebook performs best on others, suggesting complementary high- and low-level behavioral signals.

  \section{Conclusion}
  % We brought grounded theory into ML as a scalable, inductive methodology for analyzing LLM agent behavior, and introduced \framework, a multi-agent pipeline that operationalizes open, axial, and theoretical coding with an auditable trail from trajectories to theory. Across 7{,}500+ trajectories spanning six environments and four backbone LLMs, \framework reaches saturation, produces codebooks that are reproducible across runs and stable across backbones, covers 70--93\% of failure modes in independently constructed human taxonomies while surfacing patterns those taxonomies miss, recovers prior expert theoretical accounts, and yields a deductive feature space that outperforms few-shot LLM baselines on downstream failure prediction. We see the contribution as scaling grounded theory to thousands of trajectories while preserving its epistemic commitments, opening the door to applying it to richer behavioral surfaces and to using its artifacts for behavior-aware training and evaluation.

We bring grounded theory as a scalable, inductive method for studying LLM agent behavior, and build \framework, a multi-agent pipeline that runs open, axial, and theoretical coding with an auditable trail from trajectories to theory. On 7{,}500+ trajectories across six environments, \framework reaches saturation, yields codebooks reproducible stably, covers 74--91\% of failure modes in human taxonomies while surfacing patterns they miss, recovers prior expert accounts, and benefits both interpretive and predictive analysis of trajectories. We scale up grounded theory analysis to thousands of trajectories while preserving its grounding and interpretability, thus enabling richer behavioral descriptions and supporting more human-centered agent evaluation.

\paragraph{Future work.} Future work could explore richer saturation criteria that account for both the density of examples within each category and the stability of relationships between categories. This would make the algorithmic stopping rule more closely match the methodological idea of saturation. It is also worth exploring how \framework can support downstream training and evaluation. For example, they can be used to construct targeted training data, identify behaviors that should be reinforced or discouraged, design behavior-aware evaluation metrics, and build process-level reward signals~\cite{setlur2025rewarding,mahmood2024designing}. %We see this as an important direction for future work.

% We bring grounded theory into ML as a scalable, inductive method for studying LLM agent behavior, and introduce \framework, a multi-agent pipeline that performs open, axial, and theoretical coding with an auditable trail from trajectories to theory. Across 7{,}500+ trajectories from six environments, \framework reaches saturation, produces reproducible codebooks, covers 74--91\% of failure modes in human taxonomies while surfacing additional patterns those taxonomies miss, and recovers mechanisms consistent with prior expert accounts. The resulting codebooks also support downstream interpretive and predictive analyses, showing that grounded-theory-style artifacts can be useful beyond qualitative description. Together, these results suggest a path toward scalable, grounded, and human-centered evaluation of agent behavior.

  \section*{Acknowledgment}
  We thank the reviewers for their valuable comments. We gratefully acknowledge support from Google.org through the Google Cloud Research Credits Program as part of the Gemma Academic Program. This work was also supported by Schmidt Sciences, and the Texas Advanced Computing Center under grant CCR25054.
  
  \paragraph{Limitations.} Our study has several limitations. First, every coding stage in \framework is performed by an LLM, so the resulting codebooks can inherit blind spots of the backbone model; our cross-LLM stability results show the recovered structure is not dominated by any single backbone but do not rule out biases shared across frontier models. Second, the evaluations against human taxonomies and the downstream feature annotation rely on an LLM as judge, so absolute coverage and prediction numbers are conditional on the judge's reading. Third, \framework issues many LLM calls per trajectory and runs to saturation rather than a fixed budget, making it suitable for offline corpus analysis rather than per-trajectory online use. Fourth, we instantiate one qualitative methodology (grounded theory) on one behavioral surface (single-agent trajectories with success/failure outcomes) and on English-language trajectories from publicly available benchmarks; transfer to other methodologies, multi-agent or dialogue surfaces, and other languages is left to future work. Finally, the algorithmic stopping criterion operationalizes theoretical saturation but is not identical to the methodological notion.

\paragraph{License and terms of use.} All benchmark trajectories used in this work (\ALFWorld, \GAIA, \WebShop, \TauBench, \GoBrowse, and \SWEAgent) are accessed under their original licenses (Apache-2.0, MIT, or research-use), and our use is consistent with their intended research purpose. Access to OpenAI models (GPT-4.1-mini, GPT-5, GPT-5-mini) is governed by OpenAI's API terms of service; the open-weight GPT-OSS-120B is used under its released model license. All of these benchmarks were released as agent-evaluation datasets, and we use them solely to analyze agent behavior on the tasks they were designed for, which matches their intended research use.

\paragraph{Potential risks.} We also note potential risks of this line of work. Because every coding stage is run by an LLM, using the resulting codebooks to draw conclusions about agent populations can amplify biases of the backbone model, so the discovered categories should be treated as hypotheses for human review rather than ground truth about agent behavior. The framework is intended for offline analysis of publicly released benchmark trajectories and does not involve private user data; applying it to logs that contain user content would require separate consent and privacy review.

  \paragraph{Use of AI Assistants.} The authors used AI assistants (Claude and GPT-series models) during the preparation of this manuscript and the supporting code. The assistants were used for prose copy-editing, LaTeX formatting, and code refactoring. All scientific content, experimental design, analysis, and conclusions are the authors' own.

  \bibliography{latex/trajectory}

\newpage
\appendix
\setcounter{tocdepth}{2}
\tableofcontents
\allowdisplaybreaks
\newpage

\section{Details of \framework}

\subsection{Detailed algorithm with explicit state}
\label{appendix:algorithm-detailed}

Algorithm~\ref{alg:multi_agent_coding} in the main paper gives a
compact view of the pipeline. Algorithm~\ref{alg:multi_agent_coding_detailed}
below restates the same procedure with explicit state: the remaining
pool $\mathcal{D}_{\text{remain}}$, the per-round trajectory batch
$\mathcal{X}_t$ and coded record set $R_t$, the chunk-level partition
$(\mathbf{m}_1,\dots,\mathbf{m}_{M_\tau})$ of each trajectory and the
segment memo $\mu$ carried across chunks, the axial output $A_t$, and
the versioned codebook $K_t$ with revision log $L_t$. The set-builder
form makes the data-flow between agents fully explicit and is the
version we implement.

\begin{algorithm*}[!t]
\caption{Automated Grounded Theory for Agent Trajectories (detailed)}
\label{alg:multi_agent_coding_detailed}
\KwIn{Trajectory pool $\mathcal{D} = \{(\tau_i, s_i)\}_{i=1}^{N}$, batch size $B$, sampling policy $\pi$, saturation threshold $\epsilon$, stability window $W$}
\KwOut{Final codebook $K_T$, theoretical account $(c^*, \mathcal{N})$}
Initialize $\mathcal{D}_{\text{remain}} \leftarrow \mathcal{D}$, $K_0 \leftarrow \varnothing$, $t \leftarrow 0$\;
\Repeat{$t \geq W$ \textbf{ and } $a_j < \epsilon,\ \text{for all } j \in \{t-W+1,\ldots,t\}$}{
    $t \leftarrow t + 1$\;
    $\mathcal{X}_t \leftarrow \pi(\mathcal{D}_{\text{remain}}, K_{t-1}, B)$ \tcp*{\textcolor{blue}{theoretical sampling}}
    $\mathcal{D}_{\text{remain}} \leftarrow \mathcal{D}_{\text{remain}} \setminus \mathcal{X}_t$\;
    $R_t \leftarrow \varnothing$\;
    \ForEach{trajectory $(\tau, s) \in \mathcal{X}_t$}{
        Partition $\tau$ into message chunks $\mathbf{m}_1, \dots, \mathbf{m}_{M_\tau}$\;
        $\mu \leftarrow \varnothing$ \tcp*{\textcolor{blue}{segment memo}}
        \For{$k = 1, \dots, M_\tau$}{
            $(\text{codes}_k, \mu) \leftarrow \OpenCode(\mathbf{m}_k, \mu)$\;
            $R_t \leftarrow R_t \cup \{(\text{codes}_k, \tau, s)\}$\;
        }
    }
    $A_t \leftarrow \AxialCode(R_t)$ \tcp*{\textcolor{blue}{categories + relations + memo}}
    $K_t \leftarrow \Manage(K_{t-1}, A_t)$ \tcp*{\textcolor{blue}{constant comparison}}
    $a_t \leftarrow \#\{\textsc{add}\text{ actions in } L_t \setminus L_{t-1}\}$ \tcp*{\textcolor{blue}{new categories this round}}
}
$T \leftarrow t$\;
$(c^*, \mathcal{N}) \leftarrow \TheoreticalCode(K_T)$\;
\Return{$K_T,\ (c^*, \mathcal{N})$}\;
\end{algorithm*}

\subsection{Details of Corpora}
\label{appendix:corpora}

Table~\ref{tab:corpora} summarizes the two trajectory corpora used in our experiments: a \emph{failure-annotated} corpus, in which each trajectory carries expert-coded failure reasoning over three single-agent environments (\ALFWorld, \GAIA, \WebShop), and an \emph{outcome-labeled} corpus, in which each trajectory carries only a binary success/failure outcome over three environments spanning customer-service tool use (\TauBench), web browsing (\GoBrowse), and software engineering (\SWEAgent).

\begin{table*}[h]
\centering

  % \resizebox{\linewidth}{!}{%
\begin{tabular}{llccc}
\toprule
Dataset & Domain & Total trajectories & \#Successful & \#Failed\\
        &        &                    & trajectories & trajectories\\
\midrule
\multicolumn{5}{l}{\emph{Failure-annotated (expert failure reasoning)}} \\
\ALFWorld & Embodied household tasks         & $100$    & $N/A$      & $100$ \\
\GAIA     & General assistant / web reasoning & $50$     & $N/A$      & $50$ \\
\WebShop  & E-commerce web navigation         & $50$     & $N/A$      & $50$ \\
\midrule
\multicolumn{5}{l}{\emph{Outcome-labeled (success/failure only)}} \\
\TauBench & Customer service / tool use      & $1980$ & $1183$ & $797$ \\
\GoBrowse & Web browsing                     & $2000$ & $728$     & $1272$ \\
\SWEAgent & Software engineering             & $2000$ & $167$     & $1833$ \\
\bottomrule
\end{tabular}%
% }
\caption{Trajectory corpora used in this work. \#Successful / \#Failed trajectories: counts by outcome label.}
\label{tab:corpora}
\end{table*}

\subsection{Coverage Stability Analysis}
\label{app:coverage-stability}

\noindent \emph{Within-configuration replicate stability.}
We define a \emph{cell} as a dataset--backend-model tuple. With 3 datasets (\SWEGym, \GoBrowse, \SWEAgent) and 4 backend models, this yields 12 cells. For each cell, we run \framework three times under a random sampling policy, producing three saturated codebooks from disjoint trajectory subsets.

We compare two distributions of codebook-pair similarities. The \emph{in-cell} distribution contains pairs from replicate runs of the same cell. The \emph{cross-cell} distribution contains pairs from different cells, spanning different domains or different backend-model inductive biases. We use the cross-cell distribution as a data-derived null for calibrating reproducibility, avoiding an arbitrary similarity cutoff. Specifically, we use the 95th percentile of the cross-cell distribution as the high-reproducibility threshold.

In-cell codebook pairs have median cosine similarity 0.929 and mean cosine similarity 0.923. The 594 cross-cell pairs have median 0.791, range 0.57--0.94, and 95th percentile 0.901. We therefore adopt 0.901 as the data-derived high-reproducibility threshold. The in-cell and cross-cell distributions differ significantly under a Mann--Whitney $U$ test ($p < 10^{-20}$), and all 12 cell means exceed the 0.901 threshold. Expanding the analysis along the sample-size axis to 20 dataset--model--sample-size groups, all 20 group means remain above the same threshold.

\noindent\emph{Cross-LLM stability on the same dataset.}
To test whether \framework extracts structure from the trajectories rather than projecting backend-LLM priors onto them, we hold the dataset fixed and compute pairwise coverage between codebooks produced by different backend LLMs. We compare this observed cross-LLM coverage against three null distributions.

The first null is \emph{cross-dataset any-model}, which pairs codebooks from different datasets with arbitrary backend-model assignments. The second is \emph{cross-dataset same-model}, which pairs codebooks from different datasets while matching the backend model, thereby absorbing model-specific inductive bias. The third is a \emph{dataset-label permutation null}, which shuffles dataset labels across all codebooks and recomputes the corresponding coverage statistic over $1{,}000$ permutations.

As shown in Figure~\ref{fig:coverage-stability:b}, for every dataset, observed cross-LLM coverage lies above all three null distributions. The permutation test yields $p < 0.001$ for each dataset. These results suggest that cross-LLM agreement is not explained by shared model priors or generic codebook similarity, but by dataset-specific behavioral structure extracted from the trajectories.

\subsection{GLM Fit Diagnositic}

Table~\ref{tab:lr_fit} reports the model fit statistics and representative coefficients. The significant likelihood-ratio test ($p < 0.001$) indicates that the codebook-derived features carry information for distinguishing successful from failed trajectories, while the $\chi^2$-based fit statistics provide an additional diagnostic check on model fit. 

\begin{table*}[t]
\centering
\begin{tabular}{lrrcccc}
\toprule
 & & & \multicolumn{2}{c}{Explanatory} & Distributional \\
\cmidrule(lr){4-5}
Dataset & $N$ & df & Nagelkerke $R^2$ ($\uparrow$) & LR $p$ ($\downarrow$) & Pearson $\chi^2/\text{df}$ $\approx\!1$ \\
\midrule
\SWEGym   & $170$ & $36$ & $0.618$ & $<.001$ & $1.809$ \\
\GoBrowse & $734$ & $49$ & $0.431$ & $<.001$ & $0.992$ \\
\SWEAgent & $396$ & $68$ & $0.483$ & $<.001$ & $1.037$ \\
\TauBench & $792$ & $30$ & $0.106$ & $<.001$ & $1.035$ \\
\bottomrule
\end{tabular}
\caption{GLM fit diagnostics on max-balanced training subsets. Arrows indicate the preferred direction: higher $R^2$, lower LR $p$, and Pearson $\chi^2/\text{df}$ closer to 1.}
\label{tab:lr_fit}
\end{table*}

\subsection{Example emergent categories}
\label{appendix:emergence}
Three emergent failure-mode categories that the open-coding codebook surfaced but the human failure-mode taxonomy does not enumerate, one per dataset. Each is a behavioral pattern at the agent's decision / action level that the human taxonomy's module $\times$ failure\_type schema cannot express.

\begin{table*}[!t]
\centering
\begin{tabular}{l p{0.25\textwidth} p{0.56\textwidth}}
\toprule
Dataset & Category & Definition \\
\midrule
ALFWorld & \textit{noncompliant inaction} & Refusing to select any admissible action when an action is required. \\
\addlinespace
GAIA & \textit{signaled-but-unrealized shifts} & Announcing a pivot or new plan that is not enacted in subsequent actions, leaving behavior unchanged. \\
\addlinespace
WebShop & \textit{mode oscillation without synthesis} & Rapidly toggling between paging and resets (or similar moves) without integrating page feedback into a new tactic. \\
\bottomrule
\end{tabular}
\caption{Example emergent categories.}
\label{tab:emergent_categories}
\end{table*}

\subsection{Reproducibility}
\label{appendix:implementation}

The main-text results are produced with a batch size of $B = 30$ trajectories per round, an open-coding chunk size of 50 messages, the GPT-series default temperature of $1$ across all coding agents, an \textsc{add}-rate saturation threshold of $\epsilon = 0.2$ enforced over two consecutive rounds, and a disjoint uniform random sampling policy over $\mathcal{D}_{\text{remain}}$. During development, we varied each of these settings and found the qualitative findings to be robust; what changes is primarily the rate at which saturation is reached. Concretely, we tried $B \in \{10, 20\}$ and observed no meaningful difference in the saturated codebooks, with smaller batches simply taking more rounds to converge. The chunk size of 50 is an engineering choice driven by the backbone's context-window budget rather than a methodological one: other values produced no qualitative effect on the resulting codes provided the chunk plus the running segment memo fit comfortably within the context window. Codebook quality was likewise insensitive to temperature within the range we explored. For the saturation threshold, smaller values of $\epsilon$ produced systematically slower convergence without changing the qualitative shape of the saturated codebook, while larger values terminated earlier at the cost of weaker saturation guarantees; we adopted $0.2$ as the smallest value that still terminated within a reasonable compute budget on all datasets. Finally, we also experimented with different sampling policies, such as stratified sampling policies along semantic clustering centers, and sampling weights proportional to distance to sampled trajectories, and these can accelerate convergence on some datasets but do not yield consistent gains across different datasets, so we report the simpler random policy for comparability.

\subsection{Additional Validation and Ablations}
\label{appendix:additional-evaluation}

\paragraph{One-pass coverage baseline.}
To isolate the effect of iterative coding, we compare \framework with a direct one-pass baseline using the same \texttt{gpt-5-mini} backbone. The baseline codes as many trajectories as fit within one context window and returns the same codebook schema as \framework. We evaluate both codebooks against the human taxonomy with the same \texttt{gpt-5} judge and the Recall, Precision, and Match metrics defined in Section~\ref{sec:coverage}. Table~\ref{tab:one-pass-coverage} shows that \framework achieves higher Recall and Match on all three datasets. This result supports a practical benefit of iteration under the tested setup, but does not cover all possible one-pass procedures.

\begin{table*}[!t]
\centering
% \small
\begin{tabular}{llrccc}
\toprule
Dataset & Method & $|\mathcal{C}|$ & Recall (\%) & Precision (\%) & Match (\%) \\
\midrule
\multirow{2}{*}{\ALFWorld}
 & \framework & $20$ & $75.0$ & $60.0$ & $82.4$ \\
 & Single pass & $10$ & $60.0$ & $70.0$ & $60.8$ \\
\midrule
\multirow{2}{*}{\GAIA}
 & \framework & $17$ & $73.7$ & $63.2$ & $58.0$ \\
 & Single pass & $12$ & $54.5$ & $41.7$ & $46.0$ \\
\midrule
\multirow{2}{*}{\WebShop}
 & \framework & $13$ & $90.9$ & $88.9$ & $87.9$ \\
 & Single pass & $10$ & $63.6$ & $70.0$ & $78.8$ \\
\bottomrule
\end{tabular}
\caption{Coverage of \framework and a direct one-pass baseline. Both methods use \texttt{gpt-5-mini} to construct the codebook and the same \texttt{gpt-5} coverage judge.}
\label{tab:one-pass-coverage}
\end{table*}

\paragraph{Human validation of the coverage judge.}
We sample 100 human-failure-mode/codebook-category pairs, balanced across judge-labeled matches and non-matches. Two graduate students with AI/ML backgrounds independently label each pair as a match or non-match using only the two category definitions, without access to the LLM judge's decision or rationale. Table~\ref{tab:human-judge-validation} reports inter-annotator and annotator--judge agreement. The annotators attain Cohen's $\kappa=0.74$ (87\% raw agreement); their respective agreement with the judge is $\kappa=0.66$ and $0.76$.

Disagreements are asymmetric. Annotators A and B relabel 13 and 9 judge-labeled non-matches as matches, respectively, but only 4 and 3 judge-labeled matches as non-matches. Among the 87 pairs on which the annotators agree, they reverse seven judge-labeled non-matches and one judge-labeled match. On this sample, the judge is therefore more conservative than permissive relative to the human labels.

\begin{table*}[!t]
\centering
\normalsize
\textbf{(A) Inter-annotator and annotator--judge agreement}

\vspace{0.3em}
\begin{tabular}{lrccrr}
\toprule
Dataset & $n$ & \multicolumn{2}{c}{Annotator--annotator} & \multicolumn{2}{c}{Annotator--judge} \\
\cmidrule(lr){3-4}\cmidrule(lr){5-6}
& & $\kappa$ & Agreement (\%) & $\kappa$ (A) & $\kappa$ (B) \\
\midrule
\ALFWorld & $36$  & $0.70$ & $86$ & $0.51$ & $0.67$ \\
\GAIA     & $28$  & $0.64$ & $82$ & $0.71$ & $0.79$ \\
\WebShop  & $36$  & $0.83$ & $92$ & $0.78$ & $0.83$ \\
Overall   & $100$ & $0.74$ & $87$ & $0.66$ & $0.76$ \\
\bottomrule
\end{tabular}

\vspace{0.8em}
\textbf{(B) Direction of annotator--judge disagreements}

\vspace{0.3em}
\begin{tabular}{lcc}
\toprule
Comparison & Human match / judge non-match & Human non-match / judge match \\
\midrule
Annotator A & $13$ & $4$ \\
Annotator B & $9$ & $3$ \\
Both annotators agree ($n=87$) & $7$ & $1$ \\
\bottomrule
\end{tabular}
\caption{Human validation of the LLM coverage judge on 100 category pairs. Panel A reports Cohen's $\kappa$ and raw agreement; Panel B gives the direction of disagreements.}
\label{tab:human-judge-validation}
\end{table*}

\paragraph{Human audit of unmatched categories.}
The same two annotators independently review all 17 \framework categories unmatched to the human taxonomy. Using each category's definition, trajectory excerpts, surrounding context, and coding memos, they label each category as a valid failure-related behavior, a valid but failure-neutral behavior, or a spurious category. Both annotators judge all 17 categories to be valid, with complete agreement on the valid-versus-spurious decision. Fourteen categories are unanimously labeled as failure-related; the remaining three capture valid behaviors correlated with failure but are not themselves failure modes. Table~\ref{tab:unmatched-category-audit} gives the per-dataset results.

\begin{table*}[!t]
\centering
\normalsize
% \resizebox{\columnwidth}{!}{%
\begin{tabular}{lcccr}
\toprule
Dataset & Total & Valid failure (A / B) & Valid neutral (A / B) & Spurious \\
% & & (A / B) & (A / B) & \\
\midrule
\ALFWorld & $7$ & $6/5$ & $1/2$ & $0$ \\
\GAIA     & $8$ & $7/8$ & $1/0$ & $0$ \\
\WebShop  & $2$ & $2/2$ & $0/0$ & $0$ \\
\bottomrule
\end{tabular}%
% }
\caption{Human audit of all 17 \framework categories unmatched to the human taxonomy. A and B denote the two annotators.}
\label{tab:unmatched-category-audit}
\end{table*}

\paragraph{Saturation and sampling robustness.}
\noindent\emph{Merge-first instruction.}
We test whether the merge-first instruction mechanically suppresses \textsc{add} by rerunning the downstream stages on \SWEAgent with \texttt{gpt-5-mini}. The modified prompt places equal burden on \textsc{add} and \textsc{merge}; all other settings remain fixed, and early stopping is disabled. The mean \textsc{add} rate changes from 0.125 to 0.129, and the run still satisfies the $\epsilon=0.20$ stopping criterion. The resulting codebook contains 26 rather than 20 categories and has cosine similarity 0.957 to the original, above both the 0.923 reseed reference and the 0.791 unrelated-codebook baseline.

\noindent\emph{Stopping threshold.}
Across the 12 canonical \SWEAgent runs, the median terminal \textsc{add} rate is 0.056; 9 of 12 runs terminate below 0.10, and the maximum is 0.167. Tightening the threshold to 0.10 requires \texttt{gpt-4.1-mini} to process an average of 7.0 additional batches before convergence. The similarity curve in Figure~\ref{fig:saturation} approaches 1 by construction because it is measured against the terminal codebook. We therefore interpret the trajectory, rather than its endpoint, as evidence of gradual stabilization.

\noindent\emph{Sampling policy.}
On \SWEAgent, we compare uniform random sampling, stratified sampling by semantic clusters, weighted sampling of representative instances within clusters, and codebook-conditioned sampling. We use three seeds for each policy. No policy consistently dominates. Uniform random sampling reaches saturation in the fewest batches while producing codebooks of comparable final size. Codebook-conditioned sampling attains approximately 0.99 Chao1 coverage and produces fewer low-frequency categories, but converges more slowly. We therefore retain uniform random sampling as a simple and competitive default.

\paragraph{Cross-family ablation.}
The four principal coding backbones belong to the GPT family. Agreement among them therefore tests within-family stability but cannot exclude shared family-level or frontier-model biases. We run a cross-family ablation on \SWEAgent with Gemini-3-Flash. The runs converge after a median of four batches, with a mean per-run \textsc{add} rate of 0.15 and a median final codebook size of 11 categories. This ablation extends the evidence beyond the GPT family, but one model on one dataset cannot establish broad cross-family robustness.

\paragraph{Prompt-sensitivity ablation.}
We vary only the open-coding system prompt for \texttt{gpt-5-mini} on \ALFWorld. V0 is the main-experiment prompt, V1 reorders and rephrases the instructions and changes the worked examples, and V2 is a shorter rewrite. We run each variant with two trajectory orders. All six runs reach saturation after four codebook versions and produce 16--20 categories. Table~\ref{tab:prompt-sensitivity} shows that across-prompt similarity is comparable to the variation induced by trajectory order and exceeds both the 0.901 high-reproducibility threshold and the cross-dataset null.

\begin{table}[!t]
\centering
% \small
\resizebox{\columnwidth}{!}{%
\begin{tabular}{lr}
\toprule
Comparison & Mean cosine similarity \\
\midrule
Different prompts & $0.932$ \\
Same prompt,  & \multirow{2}{*}{$0.929$}\\
different trajectory orders& \\
Cross-dataset null & $0.777$ \\
Paper Q2 reseed reference & $0.923$ \\
\bottomrule
\end{tabular}%
}
\caption{Prompt-sensitivity ablation on \ALFWorld with \texttt{gpt-5-mini}. Similarity uses the codebook-centroid metric with \texttt{text-embedding-3-large}.}
\label{tab:prompt-sensitivity}
\end{table}

\begin{table*}[!t]
\centering
% \small
\begin{tabular}{lcccccc}
\toprule
& \multicolumn{2}{c}{\TauBench} & \multicolumn{2}{c}{\SWEAgent} & \multicolumn{2}{c}{\GoBrowse} \\
\cmidrule(lr){2-3}\cmidrule(lr){4-5}\cmidrule(lr){6-7}
Feature & MCC & ROC AUC & MCC & ROC AUC & MCC & ROC AUC \\
\midrule
Semantic clustering ($K=10$) & $0.268$ & $0.687$ & $0.305$ & $0.767$ & $0.180$ & $0.620$ \\
\bottomrule
\end{tabular}
\caption{Semantic-clustering baseline with the downstream classifier held fixed. Cluster membership is supplied to the same AutoML pipeline used for the codebook features.}
\label{tab:semantic-clustering-baseline}
\end{table*}

\paragraph{Semantic-clustering baseline.}
To hold the downstream classifier fixed, we embed each trajectory prefix with \texttt{text-embedding-3-large}, apply $K$-means clustering with $K=10$, and encode cluster membership as a one-hot feature vector. We then train the same AutoML pipeline with the same splits and labels. Table~\ref{tab:semantic-clustering-baseline} reports MCC and ROC AUC. Semantic clustering outperforms the \texttt{gpt-5-mini} \framework features only on \TauBench, underperforms them on \SWEAgent and \GoBrowse, and does not outperform the complementary feature set on any dataset.

\paragraph{Computational cost.}
OpenAI Batch records provide open-coding usage for the OpenAI-hosted models, including hidden reasoning tokens. Complete usage records are unavailable for the downstream axial-coding, codebook-management, and theoretical-coding stages, so we reconstruct their token counts from the artifacts read by each call and mark them as estimates. Table~\ref{tab:computational-cost} reports open-coding cost per trajectory and the total cost of producing one \texttt{gpt-5-mini} codebook.

Across the OpenAI-hosted models, open coding costs less than \$0.04 per trajectory in every setting and usually less than \$0.01. A complete \texttt{gpt-5-mini} codebook costs \$0.74--\$1.91 across the three datasets. For scale, eight minutes of human coding per trajectory at \$25 per hour would cost \$3.33 per trajectory. This estimate depends on the assumed annotation time and labor rate and does not represent a universal human-coding cost.

\begin{table*}[!t]
\centering
% \small
\textbf{(A) Open-coding cost and tokens per trajectory}

\vspace{0.3em}
\begin{tabular}{lcccc}
\toprule
Dataset & \texttt{gpt-4.1-mini} & \texttt{gpt-5-mini} & \texttt{gpt-5} & \texttt{gpt-oss-120b} \\
\midrule
\TauBench & \$$0.0019$ ($6\mathrm{k}$) & \$$0.0047$ ($9\mathrm{k}$) & \$$0.0314$ ($11\mathrm{k}$) & $\sim 9\mathrm{k}$ tokens$^*$ \\
\SWEAgent & \$$0.0043$ ($17\mathrm{k}$) & \$$0.0067$ ($21\mathrm{k}$) & \$$0.0399$ ($22\mathrm{k}$) & $\sim 17\mathrm{k}$ tokens$^*$ \\
\GoBrowse & \$$0.0042$ ($19\mathrm{k}$) & \$$0.0054$ ($22\mathrm{k}$) & \$$0.0305$ ($22\mathrm{k}$) & $\sim 19\mathrm{k}$ tokens$^*$ \\
\bottomrule
\end{tabular}

\vspace{0.8em}
\textbf{(B) Cost of one complete \texttt{gpt-5-mini} codebook}

\vspace{0.3em}
\begin{tabular}{lrrrrr}
\toprule
Dataset & Batches & Open-coding calls & Open-coding cost & Other-stage cost & Total cost \\
\midrule
\TauBench & $3$ & $91$  & \$$0.42$ & \$$0.32$ & \$$0.74$ \\
\SWEAgent & $6$ & $274$ & \$$1.21$ & \$$0.70$ & \$$1.91$ \\
\GoBrowse & $4$ & $120$ & \$$0.65$ & \$$0.43$ & \$$1.08$ \\
\bottomrule
\end{tabular}
\caption{Computational cost of \framework. Parenthesized values in Panel A are token counts per trajectory. $^*$Token counts are estimated from call inputs because complete usage records are unavailable. Other-stage costs in Panel B are reconstructed from the artifacts read by each call.}
\label{tab:computational-cost}
\end{table*}

\paragraph{Code-manager audit trail.}
Table~\ref{tab:code-manager-audit-trail} traces \texttt{edit application errors} across ten \SWEAgent codebook versions produced with \texttt{gpt-5-mini}. Version 1 adds the category with 9 supporting members. Two later versions extend its definition, and ten proposals are merged into it as near-duplicates or lower-level subpatterns. By version 10, the category contains 84 supporting members. The trace shows how the code manager consolidates related proposals while preserving each revision's recorded action and rationale.

\begin{table*}[!t]
\centering
\normalsize
\setlength{\tabcolsep}{3pt}
\renewcommand{\arraystretch}{0.95}
\begin{tabular}{ccp{0.3\textwidth}p{0.45\textwidth}}
\toprule
Version & Members & Code-manager action & Justification from the update log \\
\midrule
v$1$ & $9$ & {\raggedright \textsc{add} \texttt{edit application errors}\par} & Failures in which an edit is rejected or immediately breaks the code, including syntax, indentation, linting, and wrong-file errors. \\
v$2$ & $15$ & {\raggedright Definition extended; \textsc{merge} $\leftarrow$ \texttt{misapplied edits}\par} & Wrong-file and wrong-scope edits are already covered; the proposal and existing category have no clear behavioral distinction (Tests $1$ and $2$). \\
\multirow[t]{2}{*}{v$4$} & \multirow[t]{2}{*}{$40$} & {\raggedright Definition extended;\newline \textsc{merge} $\leftarrow$ \texttt{misdirected-or-}\newline\texttt{speculative-editing}\par} & Speculative or misdirected edits produce the same misapplication or syntax-failure mechanism. \\
& & {\raggedright \textsc{merge} $\leftarrow$ \texttt{syntax-and-}\newline\texttt{edit-recovery}\par} & Short syntax-recovery cycles are lower-level subpatterns rather than distinct categories. \\
v$5$ & $47$ & {\raggedright \textsc{merge} $\leftarrow$ \texttt{misapplied edits and corruption}\par} & The proposal describes the same mechanism: wrong-file, wrong-scope, or syntactically corrupting edits (Test $1$). \\
v$6$ & $55$ & {\raggedright \textsc{merge} $\leftarrow$ \texttt{fragile edit outcomes}\par} & Brittle or failed edits, including duplication, overwriting, and syntax damage, are instances of the existing category (Test $1$). \\
\multirow[t]{2}{*}{v$7$} & \multirow[t]{2}{*}{$64$} & {\raggedright \textsc{merge} $\leftarrow$ \texttt{syntax/error introduction}\par} & The proposal describes syntax or runtime errors caused by edits and is already covered by the existing category (Test $1$). \\
& & {\raggedright \textsc{merge} $\leftarrow$ \texttt{misapplied edits}\par} & This proposal describes the same edit-induced failure mechanism and is likewise covered by the existing category (Test $1$). \\
v$9$ & $77$ & {\raggedright \textsc{merge} $\leftarrow$ \texttt{editing/tool misuse errors}\par} & Mechanical or tool-mediated editing mistakes that immediately damage the code fall under the same failure mechanism (Test $1$). \\
\multirow[t]{2}{*}{v$10$} & \multirow[t]{2}{*}{$84$} & {\raggedright \textsc{merge} $\leftarrow$ \texttt{editing mistakes and duplication}\par} & Wrong-file edits and duplicate insertions are concrete instances or well-represented subtypes of the existing category (Test $1$). \\
& & {\raggedright \textsc{merge} $\leftarrow$ \texttt{syntax-breaking edits}\par} & Syntax-breaking changes are also concrete instances of the existing category (Test $1$). \\
\bottomrule
\end{tabular}
\caption{Evolution of \texttt{edit application errors} across ten codebook versions. Versions with multiple merge proposals span multiple rows; each row preserves the recorded code-manager action and justification.}
\label{tab:code-manager-audit-trail}
\end{table*}

% The code and analysis supporting our experiments are available at \url{https://github.com/ZhuoranLu/Qual-Agent-Behavior-Analysis}.

\paragraph{Implementation of failure prediction}

For the downstream failure-prediction experiments in Section~\ref{sec:detection}, we set up the task as forecasting eventual failure from a partially observed trajectory and take care to prevent label leakage at every stage. We first split each corpus into disjoint training and test trajectories at the trajectory level, induce the codebook (both \framework-induced and the few-shot baseline) on the training split only, and never expose the test trajectories to any coding agent. At annotation time, every trajectory---training and test alike---is uniformly truncated to its first $50\%$ of steps before the LLM annotator is asked to flag the presence of each codebook category and the co-occurrence of each axial-coding relationship; the same prefix-only view is used to fit the FLAML classifier on the training split and to evaluate it on the test split. Outcome labels are used only as the supervision target for the classifier and are never visible to either the coding agents or the annotation agent. This protocol ensures that the predictive signal reported in Table~\ref{tab:failure_pred} comes from behavioral patterns observable in the first half of a trajectory rather than from any portion of the trajectory that overlaps with the outcome.

\clearpage

\onecolumn

\subsection{Prompts used for LLM-as-a-Judge in Coverage Analysis}
\label{appendix:coverage:protocol}

% \begin{figure*}[!ht]
\begin{tcolorbox}[
  enhanced,
  breakable,
  colback=gray!5,
  colframe=gray!40,
  title={\small \OpenCode System},
  fonttitle=\bfseries,
  boxrule=0.4pt,
  arc=1pt,
  left=1pt,
  right=1pt,
  top=1pt,
  bottom=1pt
]

\begin{PromptBlock}
You are deciding whether two failure-mode descriptions (A and B) refer to the same underlying agent failure. 

A and B are symmetric: either could be a codebook category, an annotation taxon, or an analyst note - there is no reference side.

Mark match=true when A and B name the same specific agent move,
INCLUDING:

  - one description is the general behavioral pattern and the other is a concrete instance squarely under that pattern, in EITHER direction. The abstraction relationship is valid only when the abstract side, read on its own, would predict THIS specific mechanism - not when it is equally compatible with several unrelated mechanisms.

  - both describe the same agent move expressed in different vocabulary or framed at slightly different granularity (e.g. naming the immediate sub-step vs the broader process around it).
  
  - one is positively framed (the missing behavior) and the other is negatively framed (the failure outcome of that same behavior), but both point at the same underlying move.
  
  - the descriptions agree on the core agent action / inaction and on why that action / inaction is the failure.

Mark match=false when A and B describe distinct underlying mechanisms. The test is: would a fix that addresses A's mechanism also address B's? If not, they are different mechanisms. Surface-level overlap (shared tool, outcome, or keyword) is not enough on its own.

Two descriptions name different mechanisms when they disagree about WHICH agent-side move, decision, or cognitive posture is the cause of failure - when the locus of failure in the agent's loop, or the character of the failed move, is genuinely different in A and B.

Also mark match=false when one description is a broad umbrella covering many distinct mechanisms and the other is just one of those mechanisms. An umbrella is not the same as an abstraction. An abstraction predicts the specific instance; an umbrella merely includes it alongside other, unrelated cases. Self-check: if you can readily think of several distinct mechanisms that would fit equally well under the broader side, it is an umbrella -- match=false.

When A and B point at the same specific agent move (possibly at different abstraction levels), prefer match=true. Sharing a broad task family, domain, or co-occurring failure tendency is not enough on its own -only same move counts.

A: {a}
B: {b}

Reply with exactly one JSON object: {{"match": true|false, "rationale": "<one short sentence naming the shared mechanism if match=true, or the divergence if match=false>"}}
"""
\end{PromptBlock}

\end{tcolorbox}
% \end{figure*}

\paragraph{Main prompts used to build \framework} Here are the main prompts used for \OpenCode, \AxialCode, and \TheoreticalCode agents.
\begin{tcolorbox}[
  enhanced,
  breakable,
  colback=gray!5,
  colframe=gray!40,
  title={\small\OpenCode System},
  fonttitle=\bfseries,
  boxrule=0.4pt,
  arc=1pt,
  left=1pt,
  right=1pt,
  top=1pt,
  bottom=1pt
]

\begin{PromptBlock}
You are a qualitative researcher performing open coding on behavioral logs. Your goal is to identify and name behavioral patterns -- not to describe what happened, and not to diagnose why it happened.
---
## THE CORE DISTINCTION
A code is not a description of what happened. A code is a conceptual label for the behavioral quality of what happened. Ask yourself: what is this sequence of actions AN INSTANCE OF?
  Not: "what did the system do?"
  But: "what KIND OF behavior is this?"

Examples of the distinction:
  What happened -> "system requests the same directory three times"
  Bad code      -> "requests directory listing repeatedly"  [describes the action]
  Good code     -> "re-examines same ground without advancing"  [names the behavior]

  What happened -> "system tries approach A, fails, tries approach A again"
  Bad code      -> "retries failed action"  [describes the action]
  Good code     -> "persists without adjusting strategy"  [names the behavior]

If your code could serve as a caption for the log excerpt, it is too descriptive. A good code could appear in a taxonomy of behavioral patterns across many different kinds of systems, not just this one.

---
## CHUNKING 
Read the full log before doing anything. Then segment it into chunks and code each chunk.
A chunk is a sequence of steps that form a coherent behavioral episode. You decide the boundaries. A chunk can be 1 step or many steps.
Draw the boundary where the behavioral quality changes -- not where the step type changes, not at every request-response pair.
If several consecutive steps are all instances of the same behavioral pattern, they belong in one chunk, even if they look structurally different.

---
## FOR EACH CHUNK, PRODUCE:
  STEPS  -- which step numbers belong to this chunk (e.g., "4-7"). Use the original step numbers from the log.
  QUOTE  -- the most telling fragment, copied verbatim from the log. Maximum 3 lines. Choose the fragment that best justifies your code -- not necessarily the first or last line.
  CODE   -- a conceptual label of 2-5 words in verb + noun format.
            * Use the pattern: verb + noun (e.g., "abandon partial fix", "repeat failed approach", "narrow diagnostic scope")
            * Must name a behavioral quality, not an action type
            * Must be grounded in this specific log -- not imported from prior knowledge about how AI systems fail
            * If this code could have been written before reading this log, it is not grounded. Rewrite it.
            * Forbidden vocabulary: "hallucination", "context window", "tool call", "token", "prompt", "LLM", "model", "agent", or any AI/ML technical term
  MEMO   -- 2-4 sentences of analytic thinking.
            * Why did you draw the boundary here?
            * Why is this code the right conceptual label?
            * What is uncertain or ambiguous?
            * What would change your interpretation?

---
## CONSTRAINTS
* QUOTE and CODE capture what happened and what kind of thing it is. MEMO is where you may speculate about causes or meaning.

* Do not code the final status. It is the outcome. Code the process that produced it.

* Do not produce codes that merely name the step type: "calls a tool", "produces output", "requests listing" are not codes.

* If prior context is provided, use it only for continuity. If a behavior clearly continues from before, note it in the memo.

---
## OUTPUT FORMAT
Return exactly the structure below. Do not add any text before ===CODES=== or after ===END===.

===CODES===
[{
    "steps": "<start>-<end>",
    "quote": "<verbatim fragment>",
    "code": "<2-5 word conceptual label>",
    "memo": "<2-4 sentences>"
}]
===SKIPPED===
<one line listing skipped step numbers and a brief reason, or "none">
===SEGMENT_MEMO===
<3-4 sentences of plain prose. The arc of this segment: its shape, turning point, and how it connects to or breaks from what preceded it. Write it for the next reader. Do not summarize the codes.>
===END===
\end{PromptBlock}

\end{tcolorbox}
% \section{Open Coding Prompt}

\begin{tcolorbox}[
  enhanced,
  breakable,
  colback=gray!5,
  colframe=gray!40,
  title={\small\AxialCode System},
  fonttitle=\bfseries,
  boxrule=0.4pt,
  arc=1pt,
  left=1pt,
  right=1pt,
  top=1pt,
  bottom=1pt
]

\begin{PromptBlock}
You are a qualitative researcher performing axial coding. You have received open coding results from multiple behavioral logs. Your job is not to produce more codes. 
Your job is to find the structure that connects the codes that already exist.

---

## WHAT AXIAL CODING IS

Open coding broke the logs into labeled episodes. Axial coding asks: what do these labels have in common, and how do they relate?
You are looking for two things:

1. CATEGORIES

   A category is an abstract concept that names what several codes are all instances of. Example:
     codes: "orients by scanning", "orients by repository anatomy", "maps workspace before acting"
     category: "initial spatial orientation"
     These codes are all instances of the same behavioral pattern, appearing at different trajectories' beginnings.

   A category is not just a grouping label. It should be specific enough to distinguish itself from other categories, and abstract enough to apply beyond any single trajectory.

2. RELATIONSHIPS BETWEEN CATEGORIES

   Once you have categories, ask: how do they connect? What conditions produce a category? What does a category lead to? Does one category modify or interrupt another? You are not required to use a fixed framework. Let the relationships emerge from the data. But for each relationship you claim, you must show it in the codes.

---

## WHAT YOU ARE NOT DOING

* Do not rename or re-code individual episodes. Work with the codes as given. If a code is ambiguous, note it in a memo.
* Do not impose a structure you decided before reading. If the codes do not support a relationship, do not claim it.
* Do not produce a category for every code. Some codes may be outliers, one-offs, or not yet theoretically developed. Name these explicitly as UNDERDEVELOPED rather than forcing them into a category.
* Do not treat co-occurrence as causation. "These two codes often appear together" is an observation. Claim a causal relationship only when the memos support it.

---

## INSTRUCTIONS

Read all the open coding results below before doing anything. Pay attention to: codes, memos, and segment memos. The memos often contain the reasoning that justifies grouping decisions.
Then produce:
1. CATEGORIES -- each with:
   * A name (2-6 words, conceptual not descriptive)
   * A definition (what behavioral quality does this category capture?)
   * The member codes from the input, with their trajectory IDs
   * status_distribution: whether the category appears in failed only, resolved only, or both
   * status_differentiation: if the category appears in both, describe how the behavioral quality differs between failed and resolved instances. This is not about whether the category appears -- it is about whether what it looks like differs. If no difference is visible, write "none observed".
   * A category memo: what makes these codes instances of the same thing? What varies across instances? What is invariant?
   
2. RELATIONSHIPS -- each with:
   * The two (or more) categories involved
   * The nature of the relationship (one or two sentences)
   * Evidence: which specific codes / memos support this claim?

3. UNDERDEVELOPED CODES -- codes that do not yet fit any category, with a brief note on why and what additional data might develop them.

4. AXIAL MEMO -- your overall analytic reading of this batch.
   What structure is beginning to emerge? What is still unclear or contested? What would the next batch of data need to show to develop the theory?

---

## CONSTRAINTS

* Every category must be grounded: list the member codes by trajectory ID. A category with only one member is allowed but must be flagged as thin.
* Every relationship must cite evidence from the codes or memos. Do not claim relationships from general knowledge about how systems fail.
* Forbidden vocabulary in category names and relationship descriptions: "hallucination", "context window", "token", "prompt", "LLM", "model", "agent", or any AI/ML technical term.
* The status field (failed/resolved) is available to you. Use status_distribution to record where each category appears. Use status_differentiation to record how it behaves differently across statuses. Do not treat status as an explanation -- treat it as a pattern to explain.

---

## OUTPUT FORMAT

Return exactly the structure below. Do not add any text before ===CATEGORIES=== or after ===END===.

===CATEGORIES===
[
 {  
    "category_name": "<2-6 word conceptual label>",
    "definition": "<one sentence: what behavioral quality does this capture?>",
    "member_codes": [ {"trajectory_id": "<instance_id>", "code": "<exact code from input>"} ],
    "status_distribution": "<failed only | resolved only | both>",
    "status_differentiation": "<if both: how does the behavioral quality differ between failed and resolved instances? If not both: n/a>",
    "thin": <true if only one member, false otherwise>,
    "category_memo": "<3-5 sentences: what is invariant across members, what varies, what is uncertain>"
  }
]
===RELATIONSHIPS===
[
  {
    "categories": ["<category A>", "<category B>"],
    "relationship": "<1-2 sentences describing the nature of the connection>",
    "evidence": "<which specific codes or memo passages support this?>"
  }
]
===UNDERDEVELOPED===
[
  { 
    "trajectory_id": "<instance_id>",
    "code": "<exact code>",
    "note": "<why it does not fit any category, what data would develop it>" 
  } 
]
===AXIAL_MEMO===
<4-6 sentences. What structure is emerging? What is contested or unclear? What would the next batch need to show?>
===END===
\end{PromptBlock}

\end{tcolorbox}

\begin{tcolorbox}[
  enhanced,
  breakable,
  colback=gray!5,
  colframe=gray!40,
  title={\small \TheoreticalCode System},
  fonttitle=\bfseries,
  boxrule=0.4pt,
  arc=1pt,
  left=1pt,
  right=1pt,
  top=1pt,
  bottom=1pt
]

\begin{PromptBlock}
You are a qualitative researcher performing selective coding. This is the final stage of a grounded theory analysis. 
You have received a mature codebook and the update log that records how the theory evolved across batches.

Your job is to produce a theoretical account of why these trajectories fail -- not a summary of what was found, but an explanation that could not have been written before the analysis.

---

## WHAT SELECTIVE CODING IS

Selective coding does three things:

1. IDENTIFY THE CORE CATEGORY
   
   One category, or a new concept that subsumes several, that sits at the center of the phenomenon. The core category should:
   
   * Have the most connections to other categories
   * Appear across the widest range of trajectories
   * Be abstract enough to anchor a theoretical statement
   * Explain not just what happens but what governs whether a trajectory succeeds or fails

   The core category may be an existing codebook category, or it may be a new concept that names something the existing categories collectively point toward but do not individually capture.
   
2. INTEGRATE THE OTHER CATEGORIES
   Show how every other category relates to the core -- as a precondition, a consequence, a modifier, or an interruption. Not all categories will have equal weight. Some will be central; others will be peripheral or conditional.

   For each category, state:
   * Its role relative to the core (precondition / consequence / modifier / interrupter / alternative path)
   * Whether it appears in failed trajectories, resolved trajectories, or both -- and what that distribution means theoretically
   
3. WRITE THE THEORETICAL NARRATIVE
   A prose account of the failure mechanism. This is not a list of findings. It is an argument: given what this data shows, here is what determines whether a trajectory succeeds or fails, and why.

   The narrative should be falsifiable in principle -- it should make claims specific enough that a new trajectory could contradict them.

---

## WHAT YOU ARE NOT DOING

* Do not summarize the codebook category by category.

* Do not list findings. Argue from them.

* Do not use the final execution status as an explanation.
  Status is what you are trying to explain, not the explanation itself.

* Do not introduce concepts not grounded in the codebook or update log.

* Do not produce a core category that is just the most frequent category. Frequency is not theoretical centrality.

---

## INSTRUCTIONS

Read the full codebook and update log before doing anything. Pay attention to:

* Which categories are failed only vs both

* Which relationships have the highest batch_confirmed count

* Which open questions were never resolved

* How categories were merged, split, or flagged across batches --
  this evolution often reveals where the theory is most alive

Then produce:
1. CORE CATEGORY -- name, definition, and a justification for why this is the center of the phenomenon rather than another category

2. CATEGORY INTEGRATION -- for each codebook category, its role relative to the core and what its status distribution means

3. THEORETICAL NARRATIVE -- 6-10 sentences of prose that states the theory. Write it so that someone who has not seen the data could understand the failure mechanism and use it to generate predictions about new trajectories.

4. UNRESOLVED TENSIONS -- open questions or contradictions in the codebook that the theory cannot yet account for. These are not failures of the analysis; they are its honest boundaries.

---

## CONSTRAINTS

* The core category must be justified by its connections, not its frequency. Show which relationships make it central.

* Every claim in the theoretical narrative must be traceable to at least one codebook category or relationship.

* Forbidden vocabulary: "hallucination", "context window", "token",
  "prompt", "LLM", "model", "agent", or any AI/ML technical term.

* The narrative must name the failure mechanism, not just describe the failure pattern.

---
## OUTPUT FORMAT
Return exactly the structure below.
Do not add any text before ===CORE_CATEGORY=== or after ===END===.
===CORE_CATEGORY===
{
"name": "<2-6 words>",
  "definition": "<one sentence: what does this concept capture that no single existing category captures alone?>",
  "justification": "<3-5 sentences: why is this the center? Which relationships converge on it? What does it explain that other candidates cannot?>"
}
===CATEGORY_INTEGRATION===
[{
"category_name": "<existing codebook category name>",
    "role": "<precondition | consequence | modifier | interrupter | alternative path>",
    "relation_to_core": "<1-2 sentences>",
    "status_significance": "<what does its status distribution mean theoretically?>"
}]
===THEORETICAL_NARRATIVE===
<6-10 sentences of plain prose. State the failure mechanism as a theory. Make claims specific enough to be contradicted by new data. Do not use lists or headers inside this section.>
===UNRESOLVED_TENSIONS===
["<one sentence per unresolved question or contradiction>"]
===END===
\end{PromptBlock}

\end{tcolorbox}

\end{document}